\documentclass[preprint,12pt,authoryear]{elsarticle}
\usepackage{amsmath,amssymb,amsfonts}
\usepackage{algorithmic}
\usepackage{graphicx}
\usepackage{textcomp}
\usepackage{framed,multirow}
\usepackage{mwe}
\usepackage{array}
\usepackage{float}
\usepackage{comment}
\usepackage{xcolor}
\usepackage{multirow}
\usepackage{amsmath}
\usepackage{url}
\usepackage{cuted}
\usepackage{hyperref}
\usepackage[ruled,vlined]{algorithm2e}
\usepackage{adjustbox}
\usepackage{multirow}
\usepackage{caption}
\usepackage{breqn}
\usepackage{enumitem}
\usepackage{lipsum,multicol}
\usepackage{mathtools}
\usepackage{subcaption}
\usepackage{wrapfig}
\usepackage{lscape}
\usepackage{rotating}
\usepackage{tabularx}
\usepackage{float}

\usepackage{etoolbox}
\patchcmd{\pprintMaketitle}
 {\ifvoid\absbox\else\unvbox\absbox\par\vskip10pt\fi}
 {\ifvoid\absbox\else\clearpage\unvbox\absbox\par\vskip30pt\fi}
 {}{}
\patchcmd{\pprintMaketitle}
 {\hrule\vskip12pt}
 {}
 {}{}
\patchcmd{\pprintMaketitle}
 {\hrule\vskip12pt}
 {}
 {}{}
\appto{\pprintMaketitle}{\clearpage}

\usepackage[symbol]{footmisc}

\begin{document}
\begin{frontmatter}

\title{Crime scene classification from skeletal trajectory analysis in surveillance settings}

\author[address1]{Alina-Daniela Matei\corref{cor1}}
\ead{alina.d.matei10@gmail.com}
\author[address1,address2]{Estefan\'ia Talavera\corref{cor2}}
\ead{e.talaveramartinez@utwente.nl}
\author[address3]{Maya Aghaei\corref{cor2}}
\ead{maya.aghaei.gavari@nhlstenden.com}
 
\address[address1]{Bernoulli Institute, University of Groningen, Nijenborgh 9, 9747 AG, Groningen, The Netherlands}
\address[address2]{Faculty of Electrical Engineering, Mathematics and Computer Science of the University of Twente, Drienerlolaan 5, 7522 NB, Enschede, The Netherlands}
\address[address3]{Computer Vision and Data Science Professorship, NHL Stenden University, Rengerslaan 10, 8917 DD Leeuwarden, The Netherlands}

\cortext[cor1]{Corresponding author.}
\cortext[cor2]{These authors contributed equally.}

\begin{abstract}
Video anomaly analysis is a core task actively pursued in the field of computer vision, with applications extending to real-world crime detection in surveillance footage. In this work, we address the task of human-related crime classification. In our proposed approach, the human body in video frames, represented as skeletal joints trajectories, is used as the main source of exploration. First, we introduce the significance of extending the ground truth labels for HR-Crime dataset and hence, propose a supervised and unsupervised methodology to generate trajectory-level ground truth labels. Next, given the availability of the trajectory-level ground truth, we introduce a trajectory-based crime classification framework. Ablation studies are conducted with various architectures and feature fusion strategies for the representation of the human trajectories. The conducted experiments demonstrate the feasibility of the task and pave the path for further research in the field.

\end{abstract}


\end{frontmatter}

\section{Introduction}
\label{sec:introduction}

From the pattern recognition perspective, anomaly detection involves detecting events that do not conform to an expected behaviour \citep{anomaly_survey}. Nonetheless, a universal definition of what the term `anomaly' entails is hard to achieve due to the general nature of abnormal events. Given the wide range of content that the `anomaly' term can refer to, the task of anomaly detection is relevant to a multitude of areas of research: violence detection \citep{violence1}, car and traffic accident detection \citep{car1}, crime detection \citep{crime} or security screening \citep{security_screening}, among others. 

Despite the intricacies involved with anomaly detection, many efforts have been targeted towards automatizing this task due to its significant applicability in the field of forensics, especially within the video surveillance setting \citep{video_anomaly_detection_survey}\citep{ahmed2018surveillance}\citep{lee2018archcam}. Surveillance cameras capturing continuous footage of public spaces are increasingly being used to ensure public safety \citep{billion_cameras}. Nonetheless, the human monitoring capabilities of the security agencies are overwhelmed by the big volume of video surveillance data to be supervised \citep{intro_automatic_surveillance}. This leads to surveillance inefficiencies and, eventually, to undesirable situations that could have been prevented given the human monitors-cameras ratio would have been kept within manageable bounds \citep{intro_automatic_surveillance}. This human monitoring deficiency can be compensated by automating the monitoring process: ultimately, real-time anomaly detection and prediction could be employed \citep{arroyo2015expert} \citep{rabie1990applications}.

Focusing on the relevance of surveillance video data in the context of anomaly detection, one significant aspect is that this type of data captures a wide range of unrestricted and unscripted human behaviour in a variety of scenes and situations. In this ample and accessible body of data, we focus on human-related anomaly detection. This task is closely related to human activity recognition and targets the identification of abnormal activities through the analysis of human behavior \citep{human_anomaly}. 

To date, human-related anomaly detection received limited attention from the research community, which perhaps can be motivated by the lack of human-related anomaly datasets. Recently, the novel HR-Crime dataset was introduced in \citep{hr_crime}. HR-Crime, as the human-related fraction of the original UCF-Crime \citep{ucf_crime}, focuses on human-related criminal activities in surveillance videos. The availability of the novel HR-Crime dataset enables creating a benchmark for human-centered anomaly detection \citep{hr_crime}. Another notable step towards the detection of human-related anomalies is the work of \citep{mpedrnn}, which is centered on human body movements within the surveillance videos. 

In this work, we aim at human-related anomaly classification into -unevenly distributed- crime categories in HR-Crime. To this end, we build on top of \citep{mpedrnn} feature representation aiming to discover human body behavioural patterns that could uniquely identify a depicted crime. Our contributions are as follows:
\begin{enumerate}
    \item We propose a supervised as well as an unsupervised approach for expanding the HR-Crime ground truth with trajectory-level annotations. 
    \item We propose various data augmentation techniques to mitigate the dataset imbalance of HR-Crime. 
    \item We build on top of the MPED-RNN architecture \citep{mpedrnn}, introducing a crime classification framework based on the skeletal trajectory representation of human bodies in surveillance videos.
    \item We conduct experiments with varying architectures, trajectory representations and feature fusions for the task of human-related crime classification and their distinction from human normal (non-crime) behaviour.
    \item We exhaustively discuss the obtained results, also from the social science perspective, and provide insights within the context of crime scenes.
\end{enumerate}


The rest of the paper is structured as follows: Section \ref{sec:related_works} describes the related works. The dataset and our proposed pipeline for trajectory-level annotation and dataset augmentation are described in Section \ref{sec:dataset_prep}. Section \ref{sec:methodolgy} introduces the proposed methodology together with the corresponding experimental framework. In Section \ref{sec:results}, we discuss the obtained results. Finally, Section \ref{sec:conclusion} draws our final conclusions.

%
%
%
\section{Related work}
\label{sec:related_works}
A common methodology widely applied for anomaly detection is novelty detection. This involves training the detection system with what is considered as `normal' events, which sets the reference point for the later detection of particularities in the environment. The anomaly detection system is essentially trained to accurately reproduce the `normal' video data, hence yields in high errors when faced with anomalous videos \citep{novelty_1}. This approach generally entails the use of  encoder-decoder architectures \citep{enc_dec_multi_sensor}.

A high-level taxonomy of anomaly detection methodologies broadly distinguishes between traditional and deep learning methods. Traditional methods entail the extraction and analysis of features such as optical flow \citep{opticalflow1} or histograms of oriented gradients \citep{oriented_gradients} in order to detect anomalies in video settings. More recently, deep learning techniques have been vastly exploited for the task of anomaly detection \citep{deeplearning1, sparsecodingrnn}. One popular approach entails using Generative Adversarial Networks (GANs) within the novelty detection approach \citep{doshi2021online}. Works such as \citep{gans1} are used to generate internal frame representations based on a given frame and its optical flow. 
Apart from GANs, Convolutional Neural Networks (CNNs) and Long-Short Term Memory (LSTM) networks are used for features extraction (e.g. motion and visual features) and anomaly detection in \citep{lstmcnn}. Due to the visual nature of surveillance data, anomaly detection is usually approached from a visual standpoint \citep{real_world} with an emphasis on object detection \citep{any_shot} or visual attention regions in the frames \citep{visual_attention}. These approaches have been tested and benchmarked on anomaly datasets that usually capture a limited and simple range of anomalies such as people running or throwing objects, among others.

A recent approach to anomaly detection is presented in \citep{mpedrnn}, where authors investigate how human body motion patterns relate to the concepts of (ab)normality. Unlike the other works presented before, this method uses skeleton trajectories extracted for each person in a video. Skeleton coordinates capture local body movements, as well as global body motion in the scene throughout the video. The anomaly detection technique follows the novelty detection approach and the architecture used is a single-encoder-dual-decoder (i.e. one decoder reconstructs, the other is used for predictions) referred to as MPED-RNN. MPED-RNN has two structural branches: one local and one global branch which process the corresponding trajectory features and exchange information. The anomaly score of the video is computed per frame, by selecting the maximum of the reconstruction and prediction errors corresponding to the skeleton trajectories in the frame. 

Anomaly detection extends to a classification task given prior knowledge of the type of abnormalities. In this regard, the authors of \citep{ucf_crime} introduced a video surveillance dataset targeted towards the task of anomaly detection and classification, referred to as UCF-Crime. UCF-Crime consists of normal videos and videos depicting 13 anomaly categories that correspond to crime categories. UCF-Crime poses an additional challenge for the task of anomaly detection since both the `abnormal' and `normal' videos show a high degree of diversity, both in terms of environment scenes and in the complexity of events. The complexity of events is highly influenced by the number of people present in the video and the wide range of activities, which translate into high intra-class variance. 
The proposed methodology for anomaly detection on the UCF-Crime dataset given in \citep{ucf_crime} introduces a multiple instance learning (MIL) \citep{mil} approach with weakly labelled training videos. Since manually labelling each frame in a video is expensive, the authors leveraged one label of the entire video: if a video is labelled as part of an anomaly class it can be assumed that the anomaly event occurs at some point in the video. Each video, whether `normal' or `abnormal', is divided into video segments; a video is then represented by a `bag' of its segments from which C3D features \citep{c3d_feat} are extracted. A fully connected CNN is trained on the extracted features from both `normal' and `abnormal' videos. A ranking loss function is used which computes the ranking loss between the highest scored segments in the `normal' and `abnormal' bags. The aim is to train the model and adjust the loss such that higher anomaly scores are appointed to the video segments in the `abnormal' bags. 

Our proposed methodology addresses the lack of tools for anomaly classification, most specifically for human-related crime classification. We approach this by extending the work presented in \citep{mpedrnn}, with the focus on the human-related fraction of UCF-Crime, HR-Crime \citep{hr_crime}. 

%
%
%

\section{HR-Crime trajectory-level annotation}
\label{sec:dataset_prep}

\begin{figure}
    \centering
    \includegraphics[width=0.9\textwidth]{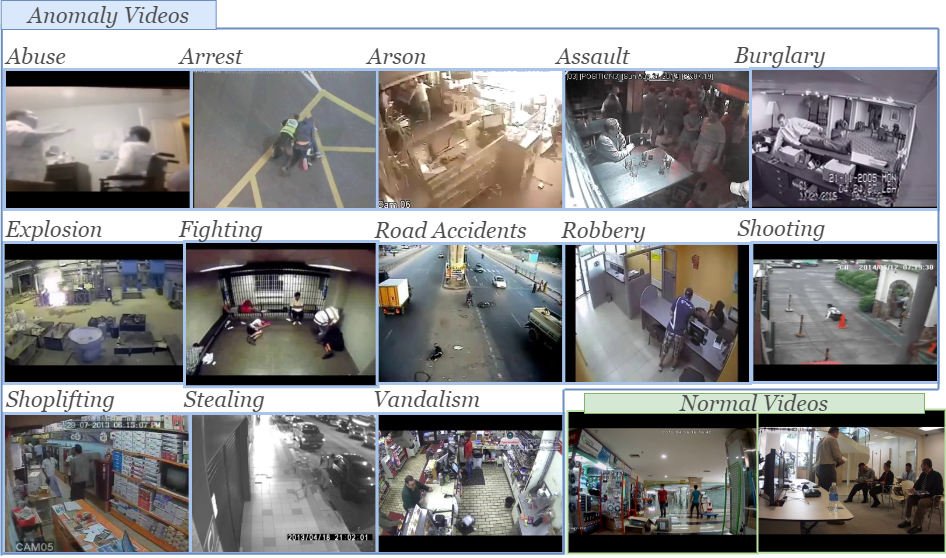}
    \caption{Example frames from HR-Crime following the class structure of the dataset.}
    \label{fig:dataset_examples}
\end{figure}

The HR-Crime dataset \citep{hr_crime}, as a subset of the UCF-Crime dataset \citep{ucf_crime} only consisting of human-related crime videos, is a significantly complex dataset due to several factors, such as low video quality and lack of content clarity even for the human eye. Sample frames extracted from the dataset are shown in Figure~\ref{fig:dataset_examples}, where the taxonomy of HR-Crime is also introduced. HR-Crime is a balanced dataset in terms of `normal' vs. `abnormal' video samples. However, as shown in Table~\ref{tab:dataset}, within the `abnormal' category, the classes are largely imbalanced.

From the ground truth perspective, each video is defined by the human skeletons and trajectories of the people appearing in it. A human skeleton is defined by 17 $(x, y)$ coordinates corresponding to 17 body joints tracked over the video. Each trajectory is split into segments of $12$ frames to ensure the requirement of models demanding fixed length inputs \citep{hr_crime}.

The HR-Crime dataset solely provides ground truth labels at video-level. Each video is annotated with the corresponding crime class or with the `normal' label for videos that do not contain an anomaly. However, it is worth noting that not all people in a crime scene are necessarily participants to the criminal activity. The video might capture bystanders or people exhibiting normal behaviour out of the abnormal event boundary. It is to be highlighted that in our approach, we aim to investigate the role of human skeletons displacement in the classification of crimes. This faces us with the lack of ground truth annotations in the HR-Crime dataset at trajectory level.

Our proposed train:test split is based on the total number of trajectories per class, rather than the number of videos. This is enabled by the novel trajectory-level ground truth introduced in Section \ref{sec:trajectory_gt}. Table~\ref{tab:dataset} shows the trajectory distribution over the 13 crime classes of HR-Crime with a 0.8:0.2 train:test ratio. The validation set is randomly set during the training phase as 0.2 of the training set.  




\begin{table}[]
    \centering
\resizebox{\textwidth}{!}{   

\begin{tabular}{l || ccccccccccccc ||cc}
Set & \rotatebox{90}{Abuse} & \rotatebox{90}{Arrest} & \rotatebox{90}{Arson} & \rotatebox{90}{Assault} & \rotatebox{90}{Burglary} & \rotatebox{90}{Explosion} & \rotatebox{90}{Fighting} & \rotatebox{90}{\begin{tabular}[l]{@{}l@{}}Road\\ Accidents\end{tabular} } & \rotatebox{90}{Robbery} & \rotatebox{90}{Shooting} & \rotatebox{90}{Shoplifting} & \rotatebox{90}{Stealing} & \rotatebox{90}{Vandalism} & \rotatebox{90}{Abnormal} & \rotatebox{90}{Normal}  \\

\hline \hline
\multicolumn{16}{c}{Trajectory based distribution}\\
\hline
Train & 221 & 186 & 101 & 290 & 227 & 50 & 309 & 59 & 579 & 107 & 588 & 178 & 45 & 2,940 & 28,622\\
Test & 55 & 47 & 25 & 72 & 57 & 12 & 78 & 15 & 145 & 27 & 147 & 45 & 11 & 736 & 7,156 \\
\hline \hline
\multicolumn{16}{c}{Video based distribution}\\
\hline
Train & 30 & 34 & 39 & 38 & 77 & 21 & 31 & 54 & 116 & 37 & 40 & 78 & 37 & 632  & 625 \\
Test & 8 & 8 & 9 & 9 & 19 & 5 & 8 & 14 & 29 & 9 & 10 & 20 & 9 & 157 & 157\\

\end{tabular}}
\caption{
Trajectory and video based dataset distributions; the trajectory based distribution follows the trajectory-level ground truth introduced in Section \ref{sec:trajectory_gt}}
\label{tab:dataset}
\end{table}

\subsection{Trajectory-level ground truth generation}
\label{sec:trajectory_gt}

In order to compensate for the lack of ground truth annotations at trajectory level, we propose a supervised as well as an unsupervised methodology for differentiating between the `normal' vs. `abnormal' trajectories over the trajectories appearing in the HR-Crime dataset.

We employ the MPED-RNN model trained \textit{de novo} on HR-Crime, as proposed in \citep{hr_crime}, in order to assign to each trajectory an anomaly score. The anomaly score, $\alpha$, per trajectory, $s_i$, is computed based on the reconstruction of the trajectory, following Equation 16 in \citep{mpedrnn}, described below in Equation \ref{eq:anomaly_score}:
\begin{equation}
\alpha_{s_i} = \frac{\sum_{u \in S_i}L_p(u)}{|S_i|} 
\label{eq:anomaly_score}
\end{equation}
where $S_i$ represents the set of reconstructed segments that compose $s_i$, for which perceptual loss $L_p$ is computed. Perceptual loss $L_p$ is defined as the mean squared error between the original trajectory segment and its reconstruction.

We propose identifying the two clusters corresponding to the `normal' and `abnormal' trajectories given the anomaly scores $\alpha$.
Based on the obtained cluster labels, we establish a new ground truth for the anomaly class classification following the steps:
\begin{enumerate}
    \item Removing `normal' trajectories corresponding to videos labelled with one of the 13 anomaly classes based on the video-level ground truth, adding them to `normal' category.
    \item Removing from the trajectories within the `normal' video-level ground truth, the falsely generated trajectories which may resemble anomalies.
\end{enumerate}
Step 1 addresses the fact that not necessarily all the people captured in an abnormal video are actively participating in the `abnormal' activity, while Step 2 mainly captures outlier trajectories in the `normal' cluster. By removing these mismatched trajectories we ensure that the HR-Crime dataset is consistent for tasks that are trajectory-related, reducing the amount of noise in the data\footnote{The trajectory-level ground truth will be publicly available upon the publication of the manuscript.}. 

As mentioned earlier, we employ a supervised and an unsupervised method to distinguish the wrongly labelled trajectories due to the video label propagation into trajectories. On one side, Gaussian mixture model \citep{gmm} clustering is used. GMM is used with two mixture components corresponding to the `normal' and `abnormal' categories with their own general covariance matrices. GMM is implemented using an Expectation-maximization probabilistic algorithm (EM) \citep{gmm} with 100 iterations and convergence threshold of $1e-3$.

\begin{figure}
    \centering
    \includegraphics[width = 0.7 \textwidth]{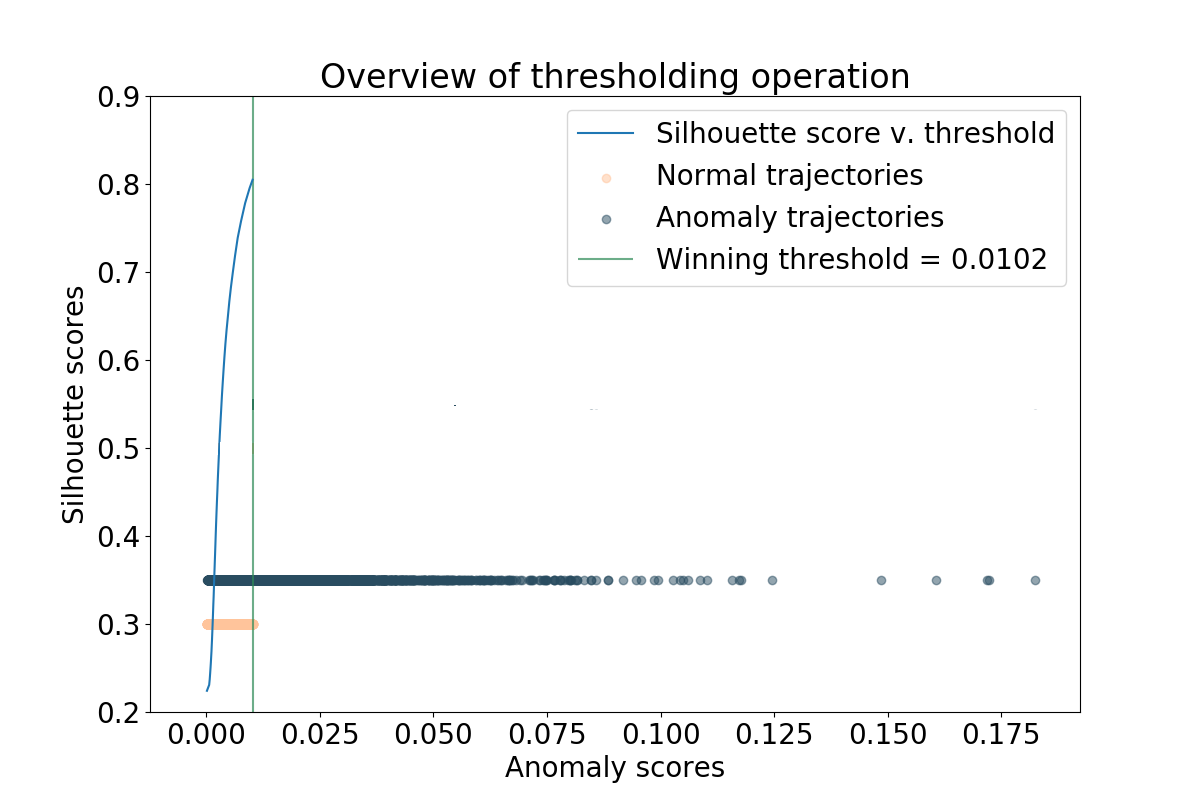}
    \caption{Depiction of the identification process for the proposed threshold in the supervised paradigm. The anomaly scores of the trajectories are plotted before applying the threshold (bottom) based on the video-level ground truth and after applying the threshold (top) based on the trajectory-level ground truth.
    }
    \label{fig:thresholding}
\end{figure}

As per the supervised approach, we plot next to each other the anomaly score of all the `normal' and `abnormal' trajectories in HR-Crime as determined by the original ground truth at video-level. We select a threshold where visually a difference in value can be spotted. We also define few more thresholds in that proximity. For each value of the threshold, we compute the Silhouette score of the resulting `normal' and `abnormal' trajectory clusters. We select the threshold which leads to the highest Silhouette score as the threshold value to separate between `normal' and `abnormal' trajectories. The final threshold is equal to the anomaly score of $0.0102$. The entire process is depicted in Figure~\ref{fig:thresholding}.

We evaluate the quality of the obtained clusters of `normal' and `abnormal' trajectories based on the Silhouette score. The obtained results of using either supervised or unsupervised technique are 0.796 and 0.752, respectively. This relatively high score indicates the decent quality of the clusters, i.e. the coherence among the grouped samples. We also evaluate the number of trajectories that received a different trajectory-level label than the video-level label provided by HR-Crime: 11,987 and 11,748 trajectories originally belonged to the `abnormal' class are relabelled as `normal', using supervised and unsupervised approach, respectively. Furthermore, 24,533 and 24,549 trajectories originally labelled as `normal' are found to be outliers that resemble the `abnormal' class using supervised and unsupervised approach, respectively.

\begin{figure}[h]
    \centering
    \includegraphics[width = 0.85\textwidth]{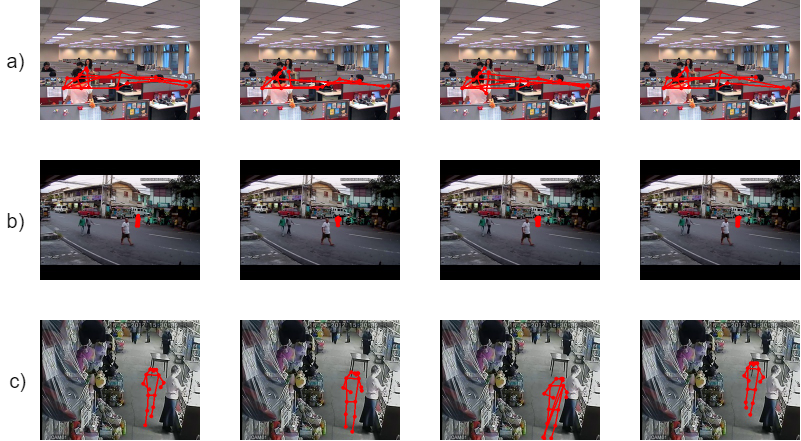}
    \caption{Examples of video-level `normal' trajectories clustered as `abnormal' within the normal category, detected by the unsupervised approach. a) skeleton merged from different skeleton joints, b) skeleton is too distant and joints overlap, c) skeleton joints not connected to an actual person are detected incorrectly.}
    \label{fig:normal_to_anomaly}
\end{figure}

Qualitatively, the results of the GMM categorization of `normal' vs. `abnormal' trajectories can be seen in Figure~\ref{fig:normal_to_anomaly}. This shows some examples of trajectories detected in `normal' video-level labelled videos that are categorized as `anomaly'. It can be seen that these trajectories do not necessarily capture anomalous crime activities, but are outliers within the HR-Crime dataset \citep{hr_crime}. 

\subsection{Dataset augmentation}
\label{sec:dataset_augmentation}
In order to alleviate the issue of class imbalance of the HR-Crime dataset, we opt for the following data augmentation techniques.

\begin{figure}
\begin{subfigure}{.5\textwidth}
  \centering
    \includegraphics[width = \textwidth]{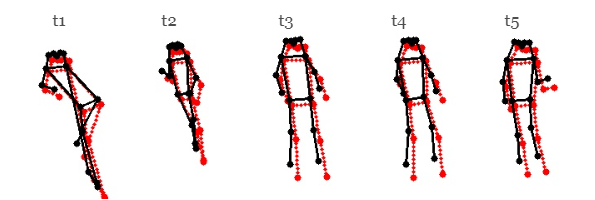}
    \caption{The original and `shifted' skeleton trajectories.}
    \label{fig:shift} 
\end{subfigure}
\hfill
\begin{subfigure}{.5\textwidth}
  \centering
    \includegraphics[width = \textwidth]{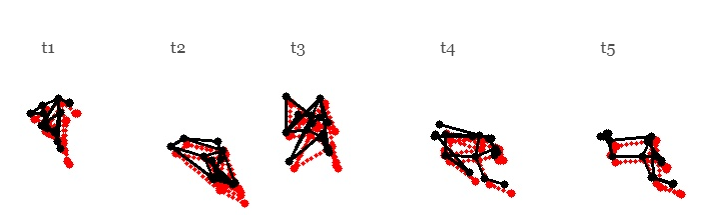}
    \caption{The noisy trajectories propagated by the skeletal `shifting' technique. }
    \label{fig:shift_noise}
\end{subfigure}
\caption{Illustration of the skeletal joints coordinate shift; `shifted' skeleton is shown in red.}
\end{figure}

\textbf{Skeletal joints coordinate shift:} A human skeleton trajectory is defined by a collection of coordinates which identifies the movement of the body over the video frames in which the body is visible. One approach to generate new trajectories is to shift the original trajectory within the bounds of the coordinate system that describe the skeletons. The shift is implemented by aggregating a small factor, $\delta$, to the skeletal coordinates in the original trajectory, resembling the skeleton being moved within the frame by a small distance. 
Figure~\ref{fig:shift} shows a trajectory segment in its original format as well as the shifted one. The `shifted' trajectory can be seen as a translation of the original trajectory. 

The value of the $\delta$ is fully dataset dependent; an inappropriate $\delta$ might negatively impact the movement behaviour captured by the original trajectory and possibly leads to transforming a normal trajectory into an abnormal one or vice versa. Hence, $\delta$ needs to accommodate two important aspects: first, represent the movement behaviour of each skeleton individually, as different people might show different particularities in terms of their body movements. Second, uniquely represent each of the  $17 (x,y)$ key points that define the skeleton movement behaviour. This is important since, for instance, hand movements do not follow the same behaviour as torso movement. In order to satisfy these two constraints, $\delta$ is computed per trajectory and per body joint based on the movement coordinates of the original trajectory. This is achieved as follows: each $\delta$ is defined as the average movement of the joint; joint movement is identified as the difference in joint coordinates between two consecutive frames in the original trajectory. The original trajectory is then used to generate any number of new trajectories by appending $\delta$ (i.e. shift to right) or $-\delta$ (i.e. shift to left) to the respective joint coordinates. Note that any number of new trajectories can be generated by altering $\delta$ with some small degree of randomness. This augmentation approach is suitable for the encoded-based classification described in Section \ref{sec:encoded_based_classif}, which entails an initial feature decomposition step applied to the trajectory. A disadvantage of applying this augmentation technique is that the resulted trajectories might introduce some degree of noise, especially for the classes which have a limited number of samples in the original dataset. This occurs due to the fact that $\delta$ is adjusted by a random value for a larger number of iterations in order to achieve the target number of trajectories. 
This effect is more undesirable considering the propagation of noisy trajectories appearing in the original dataset into the augmented dataset, as illustrated in Fig. \ref{fig:shift_noise}.

\textbf{Trajectory data point oversampling:}
Overcoming the dataset imbalance is possible by augmenting the under-represented classes with synthetic samples. Here, this is achieved using the Synthetic Minority Oversampling Technique (SMOTE) \citep{smote}. SMOTE is a suitable technique for data points with consistent size, therefore it is applicable to the trajectory segments that describe HR-Crime. Considering this aspect, SMOTE is used in relation with the decoded-based classification proposed in Section \ref{sec:decoded_based_classif}. This technique proposes the oversampling of the minority classes by creating synthetic example data until all classes have a size equal to the size of the initial majority class. The majority class in HR-Crime is \textit{Shoplifting} with almost $600$ trajectories leading to $243,272$ trajectory segments of $12$ time-steps. The minority classes are oversampled to reach this same amount of trajectory segments. The SMOTE technique generates new data while minimizing the added noise due to the technique of creating new samples based on existing neighbouring data points. This approach can add more range to the data, however, depending on the nature of the initial dataset, it is prone to result in additional noise.


\section{HR-Crime trajectory-based classification}
\label{sec:methodolgy}

Our proposed approach for classification of crime categories builds upon \citep{mpedrnn}, as we aim to study whether certain anomalous human body movements correspond to a particular crime category. The proposed models explore two types of data representation of human motion: human trajectory encodings and decodings obtained by the MPED-RNN model~\citep{mpedrnn}.
\subsection{Encoded-based classification}
\label{sec:encoded_based_classif}

\begin{figure}
    \centering
    \includegraphics[width = \textwidth]{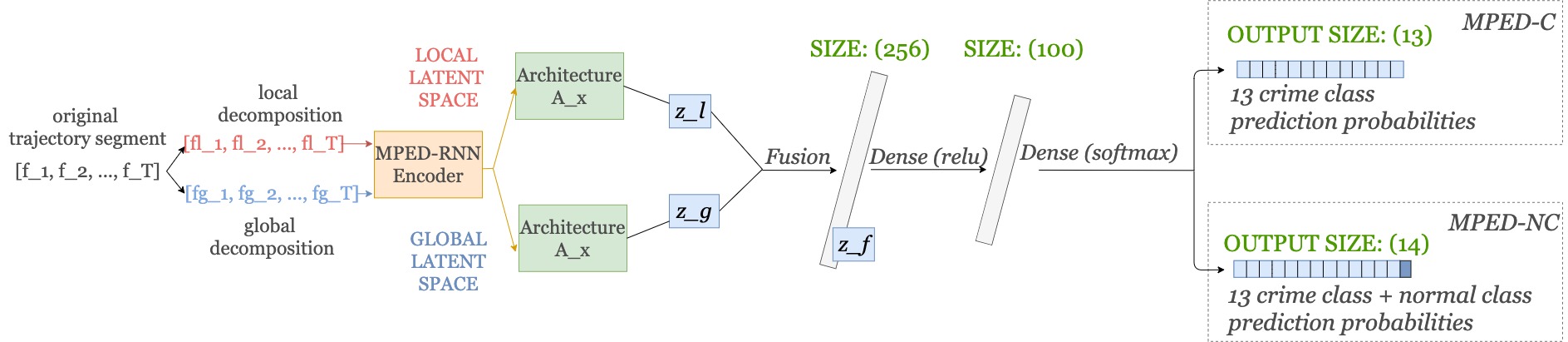}
    \caption{Illustration of the encoded-based classification models. Architecture $A_x$ refers to architectures presented in Figure~\ref{fig:end_to_end_archi}; $z_l$, $z_g$ represent the local and global feature representations which are merged into $z_f$ during the early fusion step. }
    \label{fig:mped_C_model}
\end{figure}

With this experiment, our aim is to investigate to what extent the normality encoding ability of MPED-RNN enables capturing the distinctive characteristics of each crime category. To this end, we first train MPED-RNN as per instructions given by the authors in \citep{mpedrnn} and then, isolate the trained encoder part of MPED-RNN from the rest of the architecture. Next, we freeze the isolated encoder layers in order to preserve the gained knowledge. As shown in Figure~\ref{fig:mped_C_model}, the encoder becomes the base element of the encoded-based crime classification. The local and global latent spaces resulted from the encoding process denote the feature representation corresponding to the input trajectory segment. These representations are further processed in order to reach the final classification. An overview of the architecture can be seen in Figure~\ref{fig:mped_C_model}.

\subsubsection{MPED-C: crime activities classification}
\label{sec:methodology_mped_c}
MPED-RNN encoder architecture is a two-branch RNN structure, processing the local and global representation of skeletal trajectories. In our proposed model, MPED-C, we opt for merging the local and global latent representations obtained by the encoder in an early fusion step to create a unified representation of the trajectory segment, as per Equation \ref{eq:fusion}:
    \begin{equation}
        z_f = w_l * z_l + w_g * z_g ,
        \label{eq:fusion}
    \end{equation}
where $z_l$ and $z_g$ are the local and global representations and $w_l$ and $w_g$ their corresponding aggregation weights. The multiplication is element-wise, which results in a fused trajectory segment representation, $z_f$, of the same size as $z_l$ and $z_g$, meaning that all information captured by $z_l$ and $z_g$ is preserved. The weights $w_l$ and $w_g$ are learned to specifically fine-tune the fusion process to the dataset. We also experiment with concatenating the obtained features $z_l$ and $z_g$, such that $z_f = [z_l, z_g]$. The local and global latent representations are individually processed prior to the fusion step. The fused representation is then followed by a 13 class classifier.

\begin{figure}
    \centering
    \includegraphics[width=\textwidth]{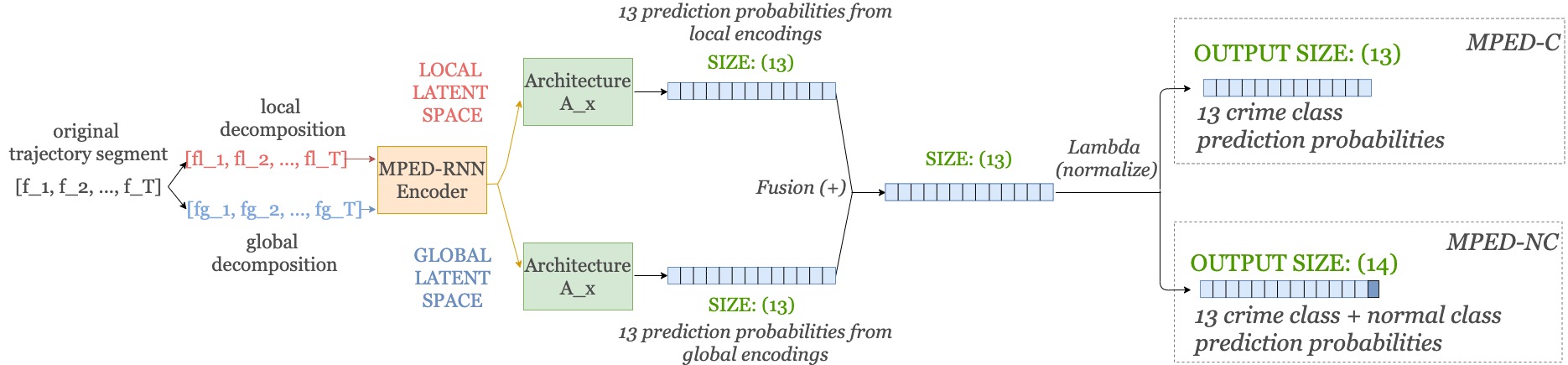}
    \caption{Illustration of late fusion of the local and global encodings; $Architecture \ A_x$ is a choice of the $A_1$, $A_2$, $A_3$ architectures visualized in Figure~\ref{fig:end_to_end_archi}.}
    \label{fig:late_fusion}
\end{figure}

\begin{figure}
    \centering
    \includegraphics[width = 0.8\textwidth]{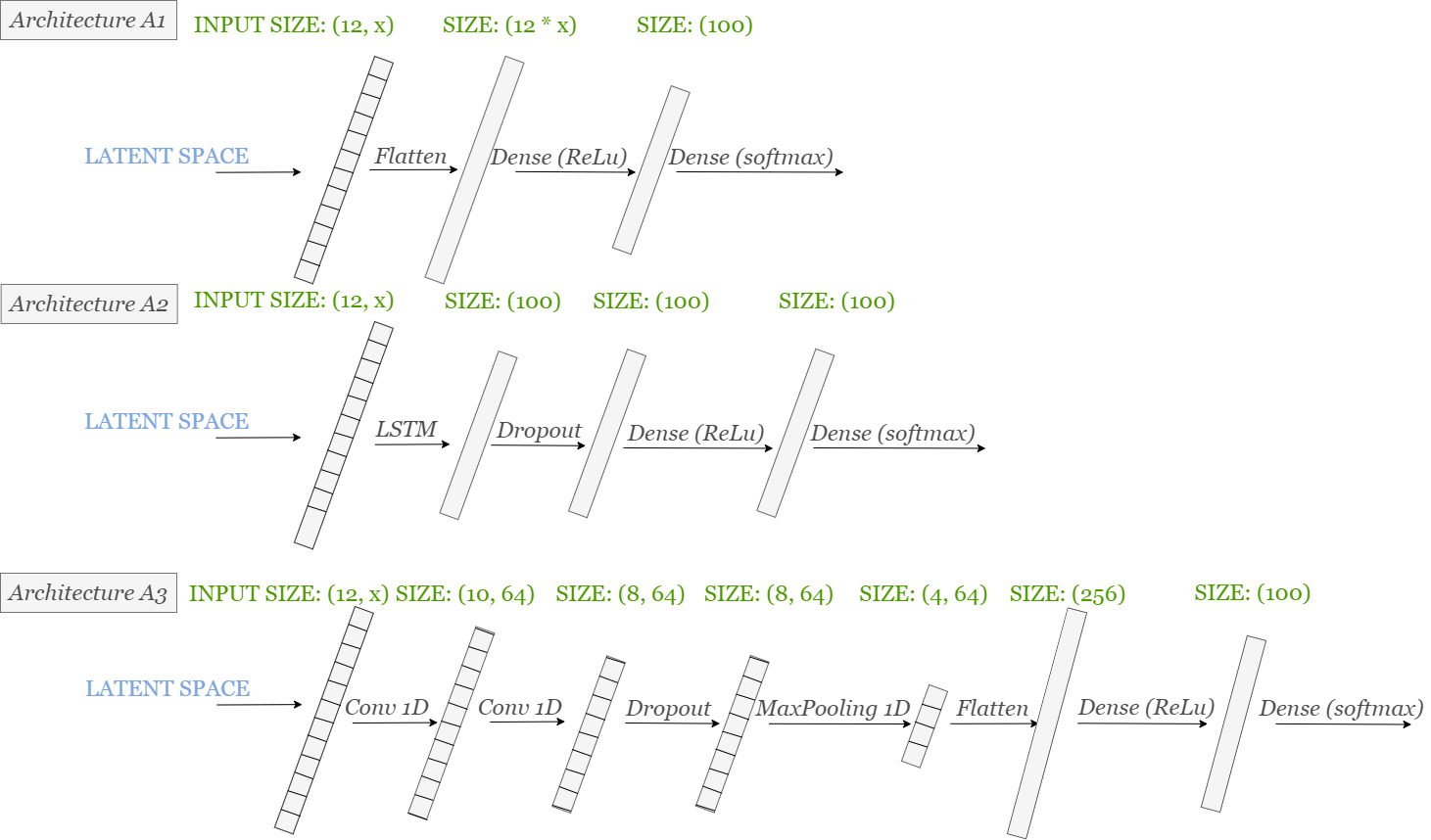}
    \caption{Visual representation of the network architectures A1, A2 and A3. The latent space refers to the MPED-RNN encoder representations.}
    \label{fig:end_to_end_archi}
\end{figure}

Furthermore, we conduct an ablation study focused on feature fusion strategies. To this end, we experiment with late fusion as follows: applying the same architecture for each of the encoding branches results into two probability vectors which yield the classification of the trajectory segments, based exclusively on their local or global features, respectively. The late fusion step involves the element-wise aggregation of the two probability vectors. The resulting vector is processed by a normalization step which gives the final classification probabilities. The key to this approach is appending the same model architecture in parallel to the local and global encoding branches. The architecture is chosen from Architectures A1, A2 and A3 shown in Figure~\ref{fig:end_to_end_archi}. An overview of this experiment is shown in Figure~\ref{fig:late_fusion}. 

We experiment with the early fusion strategy only over the best performing architecture resulting from the late fusion experiments. As we will see in Section \ref{sec:results}, the best performing architecture with late fusion is A3 and to apply early fusion with it, we discard its last two Dense layers (i.e. ReLu and softmax). Moreover, to choose the best fitting layer configuration, we conduct an ablation study for the winning architecture (i.e. A3) and the number of filters of the convolution. We run experiments with 8, 16, 32, 64, 128, 256 filters. 

\subsubsection{MPED-NC: normal and crime activities classification}
\label{sec:methodology_mped_nc}
The MPED-C model presented above is extended to recognize and classify between $14$ classes by adding the `normal' activity class to the $13$ crimes classification task. This new model is entitled MPED-NC. The adjustments required for the $14$ categories are minimal; Figures \ref{fig:mped_C_model} and \ref{fig:late_fusion} showcase the architectures of both MPED-C and MPED-NC, the only difference being the size of the last softmax activated layer with the output probability vector of $14$ instead of $13$ elements. We extend the architectural and fusion experiments presented for MPED-C in Section \ref{sec:methodology_mped_c} to the MPED-NC model.

\subsection{Decoded-based classification}
\label{sec:decoded_based_classif}
The reconstruction error is the basis of the anomaly score proposed by \citep{mpedrnn}, which essentially represents how much a trajectory deviates from `normality'. Hence, by using reconstructed trajectories as input for classifying the crimes, we implicitly capture a higher-level representation that inherently carries information about the abnormality of the trajectories in the form of the deviation of the reconstructed trajectory from the original one. With this approach, our focus is merely on the crime classification  to investigate how the degree of trajectory abnormality translates to the comprehension and distinction of the 13 crime categories, hence experiments with addition of `normal' class has been avoided.

The reconstructed trajectories by the MPED-RNN architecture are structured in a similar fashion as the original trajectories. MPED-RNN is able to reconstruct global and local features describing the skeleton, in a similar fashion to the local and global decomposition of the unprocessed skeleton; for the trajectories reconstructions, we opt for the same feature space as the original trajectories: $17 (x, y)$ pair coordinates correlated to $17$ joints of the human skeleton embedding the local and global features. The input to the crime decoded-based classifier is the reconstructed trajectories corresponding to the crime videos, structured in a $12$ time-step segment fashion.

\begin{figure}
    \centering
    \includegraphics[width = 0.75\textwidth]{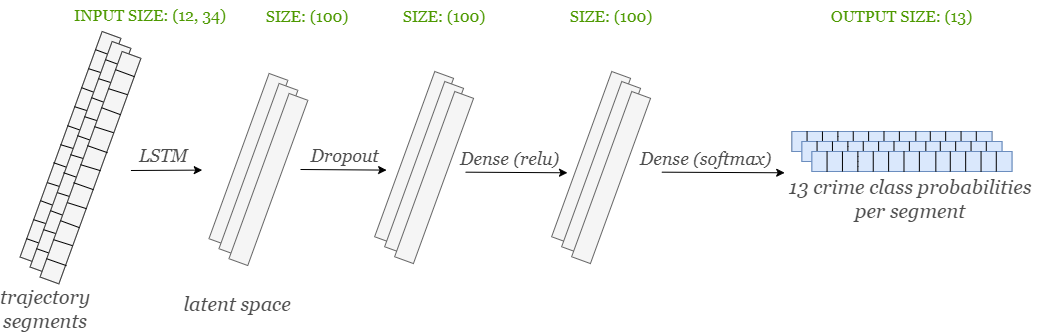}
    \caption{The proposed LSTM model for the 13 crimes classification task in Section~\ref{sec:decoded_based_classif}}
    \label{fig:13_classfier}
\end{figure}

The reconstructed skeleton trajectories have a preeminent temporal characteristic, which we assume to play a key role in solving the crime classification task. They capture a unified representation of the reconstructed body joints tracked over time, where the joints are defined by local and global features. In order to leverage the particularities of this data, we propose a classifier centered on a Long Short Term Memory (LSTM) layer \citep{lstm}. An LSTM network also acts as an encoder: it reduces the sequential input data size by reducing its dimensions to a pre-defined size, therefore creating a condensed encoding of the input data. The encodings can thereafter be manipulated further for other tasks. We simply place a fully connected layer (i.e. dense) after the LSTM for classification. The output of the proposed model is a $13$ dimension probability vector which indicates the probability of the input trajectory segment belonging to the $13$ crime classes. An overview of the proposed classifier is shown in Figure~\ref{fig:13_classfier}.

\subsection{Implementation details}
\label{sec:implementation}
The experiments are conducted with models trained using the Adam optimizer \citep{adam}, with a learning rate of $0.001$, and the categorical cross-entropy loss \citep{cross_entropy}. In order to avoid overfitting, early stopping is implemented to monitor both training and validation losses with a patience of 3 epochs.

For the extended MPED models, the training process is done conventionally using the trajectory-based split described in Section \ref{sec:dataset_prep}, since the encoder based is separately trained only on normal examples and later frozen. This implies that MPED-C, is presented with crime trajectories exclusively during training and evaluation. MPED-NC, on the other hand, is trained and evaluated both on normal and crime trajectories in order to solve the 14 class task.

For evaluation purposes, we conduct k-fold cross-validation for all the experiments discussed above. We opted for k=3 and trained the models on each fold for 25 epochs. The folds were shuffled for the training set, leaving the testing set unchanged.

\subsection{Evaluation metrics}
\label{sec:eval_metrics}

For classification evaluation, we use multi-class categorical evaluation metrics such as Accuracy, Precision, Recall and F-score. We also report the weighted average (W) over the metric performance per class based on the number of samples in the respective class, in order to account for the imbalanced nature of the original dataset. To compensate for the complexity of the classification task, we also evaluate Top-3 and Top-5 Accuracy.

Furthermore, we conduct the statistical analysis on the weighted accuracy results obtained by the models during k-fold cross-validation. The models are compared pairwise using first a Shapiro-Wilk normality test \citep{SHAPIRO1965}. If the P-value obtained is greater than $\alpha = 0.05$, a paired two-sided t-test is performed \citep{hsu2005paired}; this implies that the data comes from a normal distribution. If the P-value obtained is less than $\alpha = 0.05$, a Wilcoxon test \citep{gehan1965generalized} is used; this is a non-parametric version of the t-test, which is more suitable for data that does not come from a normal distribution.

We also present the relations between the quantitative performance of the crime classes in the form of a confusion matrix. These relations are further explored by investing the common top performance results for the class predictions and by investigating common misclassification patterns.

\section{Results and discussion}
\label{sec:results}

\subsection{Encoded-based classification results}
\label{sec:results_encoded_based}
This Section presents and discusses the results of the experiments conducted with the encoded-based classifiers introduced in Section \ref{sec:encoded_based_classif}.

\subsubsection{Convolution filter size ablation study results}
\label{sec:conv-filter}
\begin{table*}[]
    \centering
\resizebox{\textwidth}{!}{    \begin{tabular}{l||cccccc}
\#filters & Accuracy (M) & Accuracy (W) & Precision (W) & Recall (W) & F-score (W) & IBA \\
\hline\hline
8 & 0.364 $\pm$ 0.013 & 0.132 $\pm$ 0.013 & 0.328 $\pm$ 0.003 & 0.364 $\pm$ 0.013 & 0.297 $\pm$ 0.015 & 0.274 $\pm$ 0.012 \\
16 & 0.394 $\pm$ 0.006 & 0.189 $\pm$ 0.003 & 0.364 $\pm$ 0.007 & 0.394 $\pm$ 0.006 & 0.361 $\pm$ 0.003 & 0.313 $\pm$ 0.004 \\
32 & 0.419 $\pm$ 0.002 & 0.219 $\pm$ 0.002 & 0.397 $\pm$ 0.003 & 0.419 $\pm$ 0.002 & 0.395 $\pm$ 0.001 & 0.340 $\pm$ 0.002 \\
64 & 0.422 $\pm$ 0.010 & 0.239 $\pm$ 0.015 & 0.404 $\pm$ 0.015 & {0.422 $\pm$ 0.010} & {0.405 $\pm$ 0.014} & {0.343 $\pm$ 0.012} \\
128 & 0.407 $\pm$ 0.012 & 0.234 $\pm$ 0.013 & 0.393 $\pm$ 0.015 & 0.407 $\pm$ 0.012 & 0.393 $\pm$ 0.013 & 0.333 $\pm$ 0.011 \\
256 & 0.410 $\pm$ 0.017 & 0.254 $\pm$ 0.023 & 0.399 $\pm$ 0.028 & 0.410 $\pm$ 0.017 & 0.401 $\pm$ 0.023 & 0.339 $\pm$ 0.020 \\
\end{tabular}}
    \caption{Ablation study on the number of filters for the 1D Convolution layers included in the MPED-C model; training was conducted on the non-augmented version of HR-Crime.}
    \label{tab:conv_1d}
\end{table*}
As mentioned earlier, we conduct an ablation study focused on the number of filters of the convolution layers in Architecture A3 (see Figure~\ref{fig:end_to_end_archi}). The results presented in Table~\ref{tab:conv_1d} show that the model designed with 64 filters overall achieves the best performance with scores of $0.42$ and $0.24$ respectively in macro and weighted accuracy. For this ablation study, MPED-C was trained on the non-augmented version of HR-Crime. The number of filters used corresponds to the number of features that the convolution layer learns from the data; the quantitative results in Table~\ref{tab:conv_1d} show that using a limited amount of filters (e.g. 8 filters) tends to under-fit the data achieving $0.13$ weighted accuracy score, while adding excessive filters does not improve the comprehension of the data. Following these results, convolution-based architecture A3 is modelled with one-dimensional convolution layers with a filter size of 64.
\begin{table*}[!hbt]
    \centering
\resizebox{\textwidth}{!}{
\begin{tabular}{lll||ccccccc}
\begin{tabular}[c]{@{}l@{}}Model \\ ID \end{tabular} & \begin{tabular}[c]{@{}l@{}}Archi. \& \\ fusion \end{tabular}& \begin{tabular}[c]{@{}l@{}}Data \\ aug. \end{tabular}  & Accuracy (M) & Accuracy (W) & Precision (W)  & Recall (W) & F-score (W) & \begin{tabular}[c]{@{}l@{}}Top-3 \\ accuracy \end{tabular} & \begin{tabular}[c]{@{}l@{}}Top-5 \\ accuracy \end{tabular} \\
\hline \hline
\multicolumn{10}{c}{Unsupervised trajectory-level ground truth}\\
\hline
MPED-C1.1 & A1-Late & None & 0.367 $\pm$ 0.004 & 0.140 $\pm$ 0.008 & 0.310 $\pm$ 0.008 & 0.367 $\pm$ 0.004 & 0.310 $\pm$ 0.007 & 0.677 $\pm$ 0.005 & 0.824 $\pm$ 0.005 \\ 
MPED-C1.2 & A1-Late & Shift & 0.286 $\pm$	0.021 &	0.193	$\pm$ 0.004 &	0.322 $\pm$	0.014 &	0.280 $\pm$	0.021 &	0.285	$\pm$ 0.019 & 0.571 $\pm$	0.005 &	0.749 $\pm$	0.002\\ 
MPED-C2.1 & A3-Late & None & 0.403 $\pm$ 0.009 & 0.219 $\pm$ 0.018 & 0.378 $\pm$ 0.013 & 0.403 $\pm$ 0.008 & 0.380 $\pm$ 0.014 & 0.706 $\pm$ 0.007 & 0.839 $\pm$ 0.003 \\
MPED-C2.2 & A3-Late  & Shift & 0.330 $\pm$		0.020 &	0.270  $\pm$		0.010 &	0.395 $\pm$		0.012 &	0.330  $\pm$		0.020 &	0.344  $\pm$		0.018 & 0.621 $\pm$	0.017 & 0.797 $\pm$	0.010\\
MPED-C3.1 & A2-Late & None & 0.406 $\pm$ 0.002 & 0.244 $\pm$ 0.007 & 0.393 $\pm$ 0.001 & 0.406 $\pm$ 0.002 & 0.395 $\pm$ 0.002 & 0.700 $\pm$ 0.005 & 0.837 $\pm$ 0.003 \\
MPED-C3.2 & A2-Late  & Shift & 0.345	$\pm$ 0.017 &	0.271 $\pm$ 	0.017 &	0.407 $\pm$	0.017 &	0.345  $\pm$	0.017 &	0.361 $\pm$	0.017  & 0.633 $\pm$	0.013 &	0.802 $\pm$	0.001\\
MPED-C4.1 & A3-Early(c) & None & 0.423 $\pm$ 0.010 & 0.243 $\pm$ 0.012 & 0.408 $\pm$ 0.011 & 0.423 $\pm$ 0.010 & 0.408 $\pm$ 0.011 & 0.714 $\pm$ 0.007 & \textbf{0.850 $\pm$ 0.003} \\
MPED-C4.2 & A3-Early(c)  & Shift & 0.363 $\pm$	0.009 &	0.302 $\pm$	0.009 &	0.420 $\pm$	0.013 &	0.363 $\pm$	0.009 &	0.376 $\pm$	0.010 & 0.660 $\pm$ 0.003 & 0.820 $\pm$ 0.001\\
MPED-C5.1 & A3-Early(a)  & None & {0.432 $\pm$ 0.002} & 0.255 $\pm$ 0.004 & 0.418 $\pm$ 0.005 & {0.432 $\pm$ 0.002} & {0.419 $\pm$ 0.002} & \textbf{0.718 $\pm$ 0.002} & \textbf{0.850 $\pm$ 0.002} \\
MPED-C5.2 & A3-Early(a)  & Shift & 0.364	$\pm$ 0.010 &	\textbf{{0.304 $\pm$	0.008}} &{0.419 $\pm$	0.014} &	0.364 $\pm$	0.010 &	0.373 $\pm$	0.011 & 0.659 $\pm$ 0.003 &	0.816 $\pm$	0.003\\
\hline \hline
\multicolumn{10}{c}{Supervised trajectory-level ground truth}\\
\hline
MPED-C6.1 & A1-Late  & None  & 0.356 $\pm$ 0.005 & 0.140 $\pm$ 0.003 & 0.323 $\pm$ 0.006 & 0.356 $\pm$ 0.005 & 0.297 $\pm$ 0.006 & 0.655 $\pm$ 0.005 & 0.815 $\pm$ 0.005 \\
MPED-C6.2 & A1-Late  & Shift  & 0.305 $\pm$	0.004 &	0.184 $\pm$	0.006 &	0.324 $\pm$	0.007 &	0.305 $\pm$	0.004 &	0.300 $\pm$	0.004 & 0.577	$\pm$ 0.008 & 0.759 $\pm$	0.007\\
MPED-C7.1 & A3-Late  & None  & 0.413 $\pm$ 0.010 & 0.223 $\pm$ 0.012 & 0.389 $\pm$ 0.011 & 0.413 $\pm$ 0.010 & 0.388 $\pm$ 0.014 & 0.690 $\pm$ 0.012 & 0.834 $\pm$ 0.008 \\
MPED-C7.2 & A3-Late  & Shift & 0.371 $\pm$	0.029 &	0.269 $\pm$	0.025 &	0.408 $\pm$	0.033 &	0.371 $\pm$	0.029 &	0.379 $\pm$	0.031 & 0.634 $\pm$ 0.016 & 0.798 $\pm$ 0.009\\
MPED-C8.1 & A2-Late  & None & 0.418 $\pm$ 0.008 & 0.236 $\pm$ 0.005 & 0.397 $\pm$ 0.007 & 0.418 $\pm$ 0.008 & 0.398 $\pm$ 0.009 & 0.688 $\pm$ 0.007 & 0.830 $\pm$ 0.005 \\
MPED-C8.2 & A2-Late  & Shift  & 0.397 $\pm$	0.015 &	0.290 $\pm$	0.011 &	0.434 $\pm$	0.016 &	0.397 $\pm$	0.015 &	0.406 $\pm$ 0.016 & 0.657 $\pm$ 0.006 & 0.812 $\pm$ 0.002\\

MPED-C9.1 & A3-Early(c)  & None  & \textbf{0.450 $\pm$ 0.012} & 0.263 $\pm$ 0.012 & 0.434 $\pm$ 0.015 & \textbf{0.450 $\pm$ 0.012} & \textbf{0.433 $\pm$ 0.015} & 0.711 $\pm$ 0.006 & 0.840 $\pm$ 0.003 \\

MPED-C9.2 & A3-Early(c)  & Shift & 0.385 $\pm$	0.010 &	0.283 $\pm$	0.017 &	0.425 $\pm$	0.020 &	0.385 $\pm$	0.019 &	0.394 $\pm$	0.019 & 0.667 $\pm$ 0.012 & 0.818 $\pm$ 0.008\\

MPED-C10.1 & A3-Early(a)  & None  & 0.444 $\pm$ 0.015 & 0.251 $\pm$ 0.013 & 0.424 $\pm$ 0.017 & {0.444 $\pm$ 0.015} & 0.421 $\pm$ 0.015 & 0.711 $\pm$ 0.004 & 0.841 $\pm$ 0.004 \\

MPED-C10.2 & A3-Early(a)  & Shift  & 0.403 $\pm$ 0.012 & {0.291 $\pm$ 0.016} &	\textbf{0.439 $\pm$ 0.015} &	0.4031 $\pm$ 0.012	& 0.349 $\pm$	0.012 & 0.650	$\pm$ 0.015 & 0.804 $\pm$ 0.005\\

\end{tabular}}
\caption{Results of the MPED-C experiments for the classification of 13 crimes; the Architecture column refers to the different network architectures in Figure \ref{fig:end_to_end_archi} appended to the MPED-RNN encoder, fusion is either late or early (being (a) aggregation and (c) concatenation). Data augmentation is achieved by Shifting the original trajectories.}
    \label{tab:res_mped_crime}
\end{table*}

\subsubsection{MPED-C: crime activities classification results}
\label{sec:results_mped_c}
The results of the experiments conducted for the MPED-C model are presented in Table~\ref{tab:res_mped_crime}.
The best performing model is MPED-C5.2, with a performance in weighted accuracy of $0.30$.
For the remaining quantitative evaluation metrics, the best results are obtained by early fusion models, where the fusion is implemented, for most models, as aggregation. This indicates that the local and global encodings of the trajectories capture highly complementary features. Results indicate that applying the classification individually, through late fusion, does not account for the local and global inter-relation, which leads to a decrease of $0.06$ in performance. This is observed for architecture A1 in combination with late fusion. 
Comparing the performance achieved by architectures A2 and A3 in the context of late fusion, we observe that similar results are obtained; MPED-C3.2 and MPED-C2.2 both achieve aprox. $0.27$ in weighted accuracy, despite the more evolved LSTM based architecture of MPED-C3.2. 

In terms of dataset labelling, the two different resulted trajectory-level ground truths, described in Section \ref{sec:trajectory_gt}, influence the results achieved by the MPED-C experiments.
It can be observed that the top weighted accuracy is obtained by the model MPED-C5.2, trained on the trajectory-level ground truth obtained by GMM unsupervised methodology. For the remaining metrics however, top performance is achieved for models trained on the trajectory ground truth following the supervised operation. This is consistent with the results in Table~\ref{tab:res_13classifier}. Manually assuming a threshold for separating the `normal' vs. `abnormal' trajectory is customized specifically to the dataset. However, unlike the unsupervised method, it does not account for hidden relations in the data. The results also confirm the correlation between the dataset imbalance and the complexity of the 13 crimes classification task. Models trained on the augmented variant of HR-Crime, with new data samples being generated by `shifting' the original trajectories, obtain, on average, an increase of $0.04$ in weighted accuracy as compared to the same models trained on the original HR-Crime. Considering the methodology for the dataset augmentation, this circumstance might occur due to the potential added noise per class resulting from `shifting' the original trajectories, especially for those crime categories originally poorly represented. 

We report a significant increase in accuracy for the Top-3 and Top-5 accuracy metrics. Our proposed MPED-C model (with ID MPED-C5.2) obtains up to $0.82$ in Top-5 accuracy, which is significantly higher than the Top-1 accuracy of $0.36$. In general, we observe that the scores achieved in Top-3 accuracy are approximately double the macro accuracy, while Top-5 performances show even more significant increase. This indicates that the approach of representing the data through MPED-RNN encodings does not entirely overcome the challenges of low inter-class variation.
\begin{figure}[h]
    \centering
    \includegraphics[width = 0.65\textwidth]{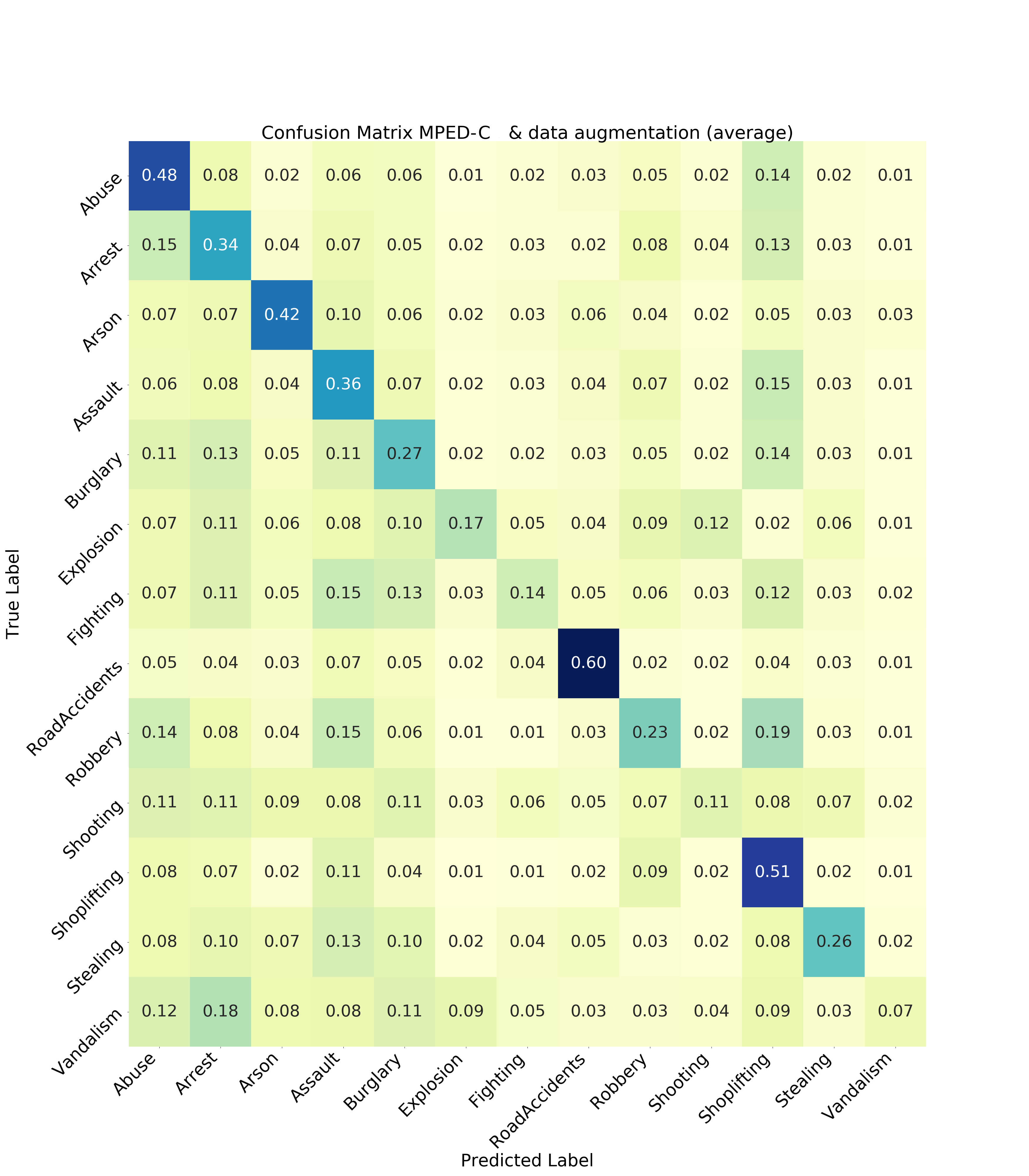}
    \caption{Confusion matrix depicting the average results using MPED-C5.2 setting.}
    \label{fig:cm_mped_crime}
\end{figure}

We apply the statistical test comparing the top performing model MPED-C5.2 with the next Top-5 best performing networks in terms of weighted accuracy: MPED-C4.2, MPED-C10.2, MPED-C8.2, MPED-C9.2, and MPED-C3.2 in Table~\ref{tab:res_mped_crime}. The obtained P-values are of 0.778, 0.083, 0.415, 0.153, 0.151, respectively. With a confidence threshold of $\alpha=0.05$, we accept the null hypothesis for all the 5 comparisons, based on the reported P-values. This similarity in achieved results can be accounted for by the fact that all compared models were trained on the augmented variant of HR-Crime.

The confusion matrix representing the results obtained by the best performing MPED-C model is presented in Figure~\ref{fig:cm_mped_crime}. As can be seen, class \textit{Road Accidents} shows the highest class accuracy of $0.60$. The achieved top result can be correlated with the distinctive characteristics of this class in comparison to the remaining 12 crimes in the dataset. \textit{Road Accidents} samples present a limited number of human participants, usually their skeletons being to some extent obstructed by their location: for example, people in cars or other vehicles. The anomaly window is also quite brief, corresponding to the moment of impact. Classes \textit{Shoplifting} and \textit{Abuse} obtain, on average, results near the $0.50$ score, with $0.51$ and $0.48$ accuracy for \textit{Shoplifting} and \textit{Abuse}, respectively. The rest of the crime classes achieve accuracy scores on the lower end with  \textit{Vandalism} obtaining the lowest average performance of $0.07$ accuracy.

Analysing the misclassification pattern of the results, we observe that the model faces some difficulties in demarcating the 13 crime categories. The misclassifications are spread across the entire confusion matrix, with classes like \textit{Abuse}, \textit{Arrest}, \textit{Assault} and \textit{Shoplifting} being most of the time wrongly predicted. The highest misclassification rates are reported for \textit{Vandalism} versus \textit{Arrest} and \textit{Robbery} versus \textit{Shoplifting} with an occurrence of aproximmately $19\%$. Class \textit{Vandalism} is also misclassified with a high frequency for classes \textit{Abuse} and \textit{Burglary}. This indicates that the model is not able to properly learn the \textit{Vandalism} category. This can be explained by the low representation of the class in the dataset, which even the augmentation procedure does not help in identifying the features of the crime category sufficiently.

The misclassification between \textit{Arrest} and \textit{Abuse} categories can be explained by the highly aggressive nature of the actions within these classes and the large overlap they share: 
the nature of the representative videos include police officials showcasing abusive behaviour. Class \textit{Fighting} frequently overlaps with classes such as \textit{Arrest}, \textit{Assault}, \textit{Shoplifting} which are characterized by increased levels of body aggression.

\begin{figure}[h]
    \centering
    \includegraphics[width=0.9\textwidth]{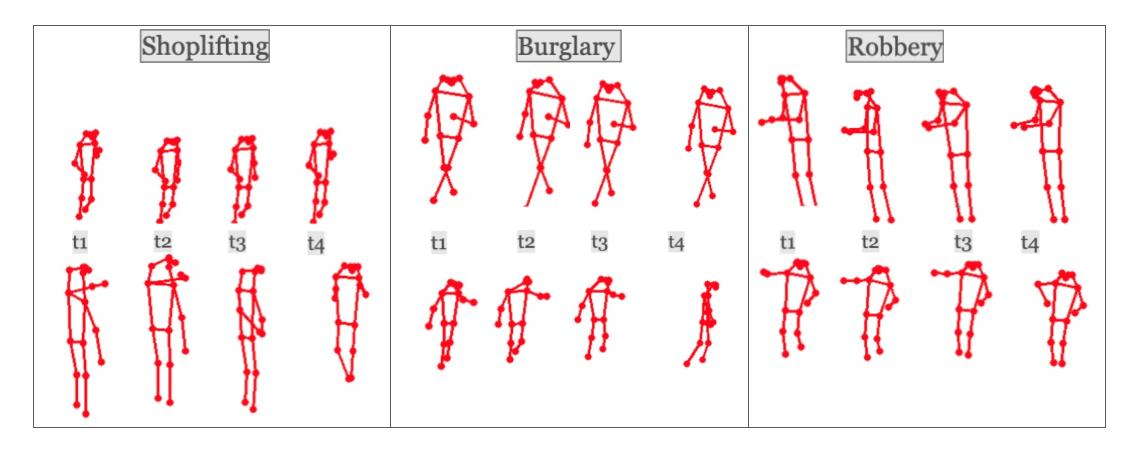}
    \caption{Trajectories relevant for crime classes \textit{Shoplifting}, \textit{Burglary} and \textit{Robbery} depicting the body movement and stance similarities shared by the three classes; the trajectories present people with a inner drawn posture inspiring a dishonest behaviour. }
    \label{fig:shoplifting_trajectories}
\end{figure}

One crime that is commonly misattributed is class \textit{Shoplifting}. This crime category has the most original trajectory samples, with 588 trajectories in the training set, therefore almost no data augmentation was required. This class also achieves the second best accuracy score of $0.54$, which can be justified by the high number of original trajectories. The most striking misclassification occurs with class \textit{Robbery}, with an average frequency of $19\%$. The similarities between the two crimes lie in the main activity in which the human participants are involved. This also explains the $15\%$ overlap with class \textit{Burglary}. The three classes are focused on the act of stealing, which implies semantic similarities. Given the common activity in which the participants are involved, human movement behaviour is shared over these categories, as showcased in Figure~\ref{fig:shoplifting_trajectories} through the exposed trajectories. \textit{Shoplifting} is focused on indoor shop environments in which one or two people are attempting to steal some object(s) without being noticed; these people usually leave the shop without any altercations. The \textit{Robbery} category showcases crimes occurring in shops where the cashier or other employees are threatened, in some cases at gunpoint, by the perpetrator in order to obtain money or some other valuable objects. The \textit{Robbery} category involves more violence than \textit{Shoplifting} which can also be deduced from its increased misclassification rates with classes \textit{Abuse} and \textit{Assault}, whereas 
\textit{Shoplifting} is misclassified as \textit{Assault} only $11\%$ of the time. However \textit{Abuse}, \textit{Arrest}, \textit{Assault} are, on average, wrongly predicted as \textit{Shoplifting} at a $0.13$ rate. This indicates that more than one label per video, and thus, per trajectory, might be necessary in order to more appropriately describe the essence of a criminal activity. 

Other categories that involve the act of stealing are \textit{Burglary} and \textit{Stealing}. \textit{Burglary} is misclassified as \textit{Shoplifting} with an average rate of $0.14$. This occurrence can be explained by the semantic nature of the class: \textit{Burglary} captures breaking and enterings in private homes or stores, with mostly one human subject involved (i.e. the perpetrator); the majority of \textit{Burglary} anomalies are captured at night-time. The overlap between \textit{Shoplifting} and \textit{Stealing} is minimal. The \textit{Stealing} category focuses on events occurring outdoors that usually involve a vehicle, either as a means of escape or, more predominantly, as the object of interest for the perpetrators. The movement behaviour is more varied and usually involves rapid, hastened local body movements describing rather short trajectories.

\subsubsection{MPED-NC: normal and crime activities classification results}
\label{sec:results_mped_nc}
The MPED-NC results are presented in Table~\ref{tab:res_mped_nc}. The highest achieving setting is the MPED-NC5, achieving a $0.24$ weighted accuracy score. This implies that for the 14 class classification task, early fusion and trajectory-level ground truth obtained using the unsupervised method achieves top classification results. Comparing the macro and weighted accuracy over all the experiments, we observe that the performance gap is relatively small; macro accuracy is on average with only $0.02$ higher than weighted accuracy, which indicates that the used dataset setting is suitable for the experiments involving 14 classes.

\begin{table*}[]
    \centering

\resizebox{\textwidth}{!}{    
   \begin{tabular}{lll||ccccccc}
Model ID  & Arch. & Fusion & Accuracy (M) & Accuracy (W) & Precision (W) & Recall (W) & F-score (W)  & \begin{tabular}[c]{@{}l@{}}Top-3 \\ accuracy \end{tabular} & \begin{tabular}[c]{@{}l@{}}Top-5 \\ accuracy \end{tabular} \\
\hline \hline
\multicolumn{10}{c}{Unsupervised trajectory-level ground truth}\\
\hline
MPED-NC1 & A1 & Late & 0.229 $\pm$ 0.011 & 0.178 $\pm$ 0.006 & 0.459 $\pm$ 0.008 & 0.229 $\pm$ 0.011 & 0.274 $\pm$ 0.014 & 0.418 $\pm$ 0.015 & 0.533 $\pm$ 0.015 \\
MPED-NC2 &A3 & Late & 0.180 $\pm$ 0.003 & 0.214 $\pm$ 0.010 & 0.473 $\pm$ 0.024 & 0.180 $\pm$ 0.003 & 0.188 $\pm$ 0.007 & 0.337 $\pm$ 0.006 & 0.442 $\pm$ 0.008 \\
MPED-NC3 &A2 & Late & 0.240 $\pm$ 0.011 & 0.239 $\pm$ 0.004 & 0.513 $\pm$ 0.003 & 0.240 $\pm$ 0.011 & 0.275 $\pm$ 0.014 & 0.411 $\pm$ 0.016 & 0.521 $\pm$ 0.012 \\
MPED-NC4 &A3 & Early(c) & 0.201 $\pm$ 0.013 & 0.224 $\pm$ 0.006 & 0.517 $\pm$ 0.021 & 0.201 $\pm$ 0.013 & 0.225 $\pm$ 0.020 & 0.381 $\pm$ 0.013 & 0.500 $\pm$ 0.014 \\
MPED-NC5 &A3 & Early(a) & \textbf{0.243 $\pm$ 0.013} & \textbf{0.244 $\pm$ 0.010} & 0.529 $\pm$ 0.010 & \textbf{0.243 $\pm$ 0.013} & \textbf{0.277 $\pm$ 0.015} & \textbf{0.450 $\pm$ 0.018} & \textbf{0.579 $\pm$ 0.021} \\
\hline\hline
\multicolumn{10}{c}{Supervised trajectory-level ground truth}\\
\hline
MPED-NC6 &A1 & Late & 0.213 $\pm$ 0.014 & 0.172 $\pm$ 0.015 & 0.450 $\pm$ 0.011 & 0.213 $\pm$ 0.014 & 0.248 $\pm$ 0.015 & 0.340 $\pm$ 0.019 & 0.517 $\pm$ 0.021 \\
MPED-NC7 &A3 & Late & 0.205 $\pm$ 0.001 & 0.218 $\pm$ 0.006 & 0.494 $\pm$ 0.012 & 0.205 $\pm$ 0.001 & 0.224 $\pm$ 0.005 & 0.370 $\pm$ 0.005 & 0.481 $\pm$ 0.009 \\
MPED-NC8 &A2 & Late & 0.230 $\pm$ 0.022 & 0.221 $\pm$ 0.019 & 0.491 $\pm$ 0.026 & 0.237 $\pm$ 0.022 & 0.271 $\pm$ 0.027 & 0.381 $\pm$ 0.016 & 0.481 $\pm$ 0.012 \\
MPED-NC9 &A3  & Early(c) & 0.204 $\pm$ 0.008 & 0.231 $\pm$ 0.012 & 0.540 $\pm$ 0.010 & 0.204 $\pm$ 0.008 & 0.221 $\pm$ 0.013 & 0.359 $\pm$ 0.008 & 0.468 $\pm$ 0.009 \\
MPED-NC10 &A3  & Early(a) & {0.218 $\pm$ 0.016} & {0.237 $\pm$ 0.011} & \textbf{0.549 $\pm$ 0.005} & {0.218 $\pm$ 0.016} & {0.243 $\pm$ 0.022} & 0.392 $\pm$ 0.028 & 0.507 $\pm$ 0.030 \\
    \end{tabular}}
    \caption{Results for the MPED-NC experiments for the classification of the 14 classes. The `Arch.' column refers to the architectures in Figure \ref{fig:end_to_end_archi} appended to the MPED-RNN encoder. (a) refers to aggregation and (c) to concatenation of features.}
    \label{tab:res_mped_nc}
\end{table*}

We can observe an overall performance drop that results from the addition of a 14th `normal' class ($0.24$ vs. $0.30$ for MPED-C in weighted accuracy). This might be due to the fact that the `normal' category shows the most intra-class variation in terms of descriptive features and trajectories. 
The complexity of the classification problem is therefore enhanced by the addition of the class since the model is faced with a 2-fold challenge: learn the inner variation of the `normal' trajectories while differentiating it from the rest of the 13 crime categories.

We apply the statistical test comparing the top performing model MPED-NC5 with the next Top-5 best performing networks in terms of weighted accuracy: MPED-NC3, MPED-NC10, MPED-NC9, MPED-NC4, and MPED-NC8 in Table~\ref{tab:res_mped_nc}. The obtained P-value is of 0.451, 0.683, 0.027, 0.198, and 0.154, respectively.
Based on the resulted P-values, we accept the null hypothesis with a confidence threshold of $\alpha = 0.05$ for all 5 comparisons. This entails that the models analysed produce consistent results, which can be accounted by the uniform data settings on which all models are trained. 

\begin{figure}[hbt]
    \centering
    \includegraphics[width=0.65\textwidth]{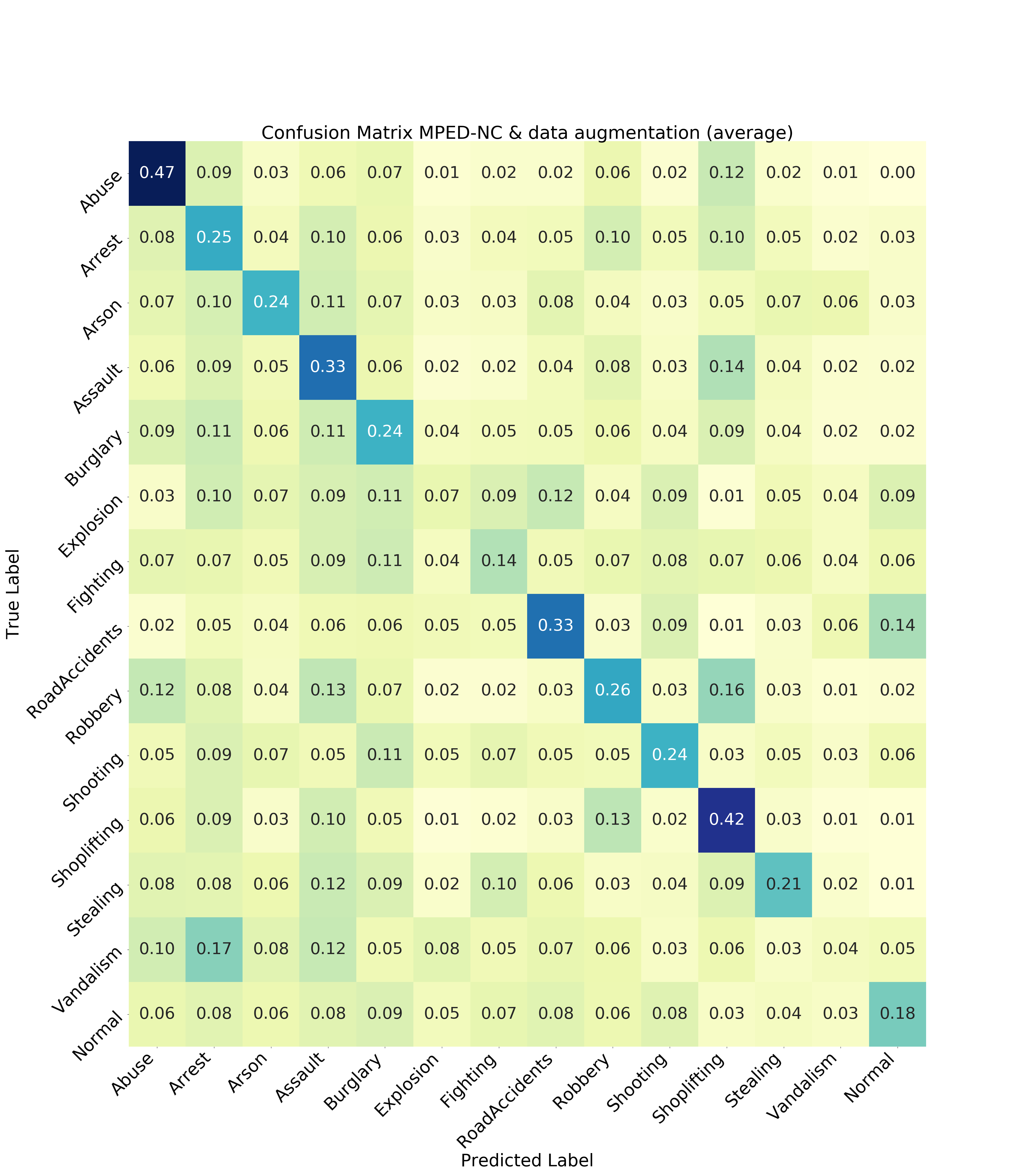}
    \caption{Confusion matrix depicting the average results using MPED-NC5 setting.}
    \label{fig:cm_mped_ncrime}
\end{figure}

\begin{figure}[hbt]
    \centering
    \includegraphics[width =0.8\textwidth]{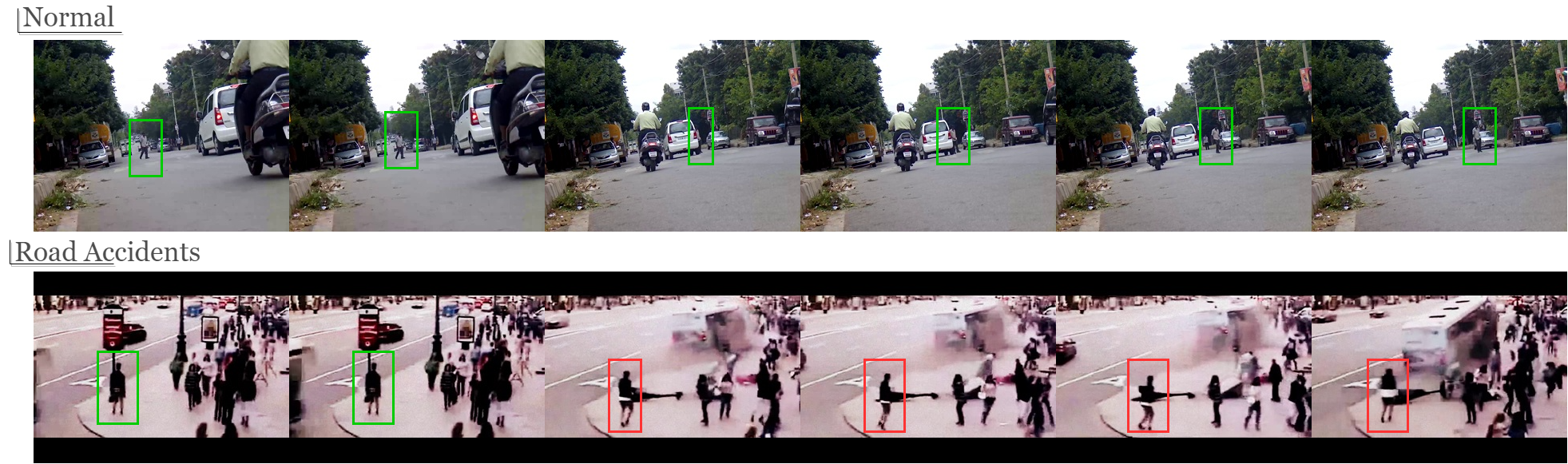}
    \caption{Depiction of two human skeleton trajectories from categories `normal' and \textit{Road Accidents}; displayed frames are not consecutive. Green boxes identify `normal' behaviour, while red boxes identify `abnormal'.}
    \label{fig:roadacc_normal_mis}
\end{figure}

The confusion matrix shown in Figure~\ref{fig:cm_mped_ncrime} displays the classification results obtained by the MPED-NC as the average over the 3-folds of the cross-validation. The top results per class are obtained for \textit{Abuse} with a score of $0.47$ and \textit{Shoplifting} with $0.42$. The remaining classes, including the `normal' class, obtain, on average, an individual accuracy of $0.24$, with no class over-passing the $0.4$ score. We exclude from this account classes \textit{Explosion} and \textit{Vandalism}, which achieve low score of about $0.07$. The \textit{Road Accidents} category achieves an accuracy score of $0.33$, while the MPED-C (see Figure~\ref{fig:cm_mped_crime}) obtained significantly higher results around the $0.6$ accuracy score. The decrease in performance follows from the notable misclassification rate of $0.14$ reported for classes \textit{Road Accidents} and the `normal' class. This can be attributed to the shared environment, the `normal' class containing videos captured by traffic cameras. Moreover, as Figure~\ref{fig:roadacc_normal_mis} shows, trajectories specific to the two classes share some commonalities. Since surveillance footage has not been pre-processed to only capture the anomaly window, situations such as the one presented in Figure~\ref{fig:roadacc_normal_mis} are frequent: the trajectory extracted from the \textit{Road Accident} example is `normal' (marked by a green box) previous to the moment of impact (i.e. the accident); afterwards, the movement behaviour anomaly (marked by a red box) becomes evident. Comparing the `normal' segment of the \textit{Road Accidents} trajectory, we observe that it shares similarities with the trajectory captured in the `normal' video: both trajectories depict pedestrians walking on the sidewalk or over a cross-walk. This observation indicates that a ground truth labelling of the trajectory segments might be beneficial for the classification since the anomaly window would be isolated, resulting in a more accurate definition of the events.

The `normal' class achieves an average of $0.18$ individual accuracy, with some misattributions occurring. This indicates that MPED-NC faces difficulties in generalizing the trained knowledge about the `normal' set. Considering the high intra-class variance as well as the random under-sampling that is applied to the `normal' class prior to training, it might be the case that the examples shown to the model during the training phase are not entirely relevant to the overall contents of the `normal' category. Revising the under-sampling step would potentially improve the per class performance for the `normal' category, given that the model would be trained on possibly more relevant examples. One way to approach this would be by providing an inner-class taxonomy and equally sampling from the `normal' sub-classes.

The frequency of misclassification observed in Figure~\ref{fig:cm_mped_ncrime} is consistent with the quantitative results presented in Table~\ref{tab:res_mped_nc}. We can observe that misclassifications occur with a higher frequency; classes such as \textit{Abuse}, \textit{Arrest}, \textit{Assault}, \textit{Burglary} and \textit{Shoplifting} are often incorrectly predicted. The classification is especially diffused between the first five classes. The most striking misclassification result is observed for \textit{Vandalism} and \textit{Arrest}, with \textit{Arrest} being predicted $17\%$ of the time for samples belonging to class \textit{Vandalism}. This might indicate that the trajectories capture two anomalous activities which can be accounted for by multi-label ground truth. The increased amount of varied misclassification is an indicator of the added complexity from including the 14th class.

\subsection{Decoded-based classification results}
\label{sec:results_decoded_based}
The results obtained by the model proposed in Section \ref{sec:decoded_based_classif} for the 13 crimes classification task, are presented in Table~\ref{tab:res_13classifier}. Model M1.2 obtains the highest performance of $0.382$ in weighted accuracy. A similar performance is obtained by the model trained on the trajectory-level ground truth after supervised thresholding; this model achieves $0.364$ in weighted accuracy on the SMOTE augmented dataset. There is an increase in weighted accuracy for the model applied on the augmented as opposed to the original dataset. Therefore, we can state that the dataset imbalance has a significant influence on the complexity of the classification task. \begin{table}[ht]
    \centering
\resizebox{\textwidth}{!}{
\begin{tabular}{lll||ccccccc}
\begin{tabular}[c]{@{}l@{}}Model \\ ID \end{tabular} & \begin{tabular}[c]{@{}l@{}}Ground \\ truth \end{tabular}  & \begin{tabular}[c]{@{}l@{}}Data \\ aug. \end{tabular}  & Accuracy (M) &  Accuracy (W) & Precision (W) & Recall (W) & F-score (W)   & \begin{tabular}[c]{@{}l@{}}Top-3 \\ accuracy \end{tabular} & \begin{tabular}[c]{@{}l@{}}Top-5 \\ accuracy \end{tabular}\\
\hline \hline
M1.1 & unsuper. & None & 0.478 $\pm$ 0.005 & 0.324 $\pm$ 0.003 & 0.477 $\pm$ 0.002 & 0.478 $\pm$ 0.005 & 0.476 $\pm$ 0.004 & 0.629 $\pm$ 0.007 & \textbf{0.861 $\pm$ 0.004} \\
M1.2 &{unsuper.} & {SMOTE} & {0.408 $\pm$ 0.007} & \textbf{0.382 $\pm$ 0.005} & { 0.497 $\pm$ 0.001} & {0.408 $\pm$ 0.007} & {0.428 $\pm$ 0.006} & 0.577 $\pm$ 0.010 & 0.846 $\pm$ 0.003 \\
\hline \hline
M2.1 & super. & None & \textbf{0.503 $\pm$ 0.005} & 0.318 $\pm$ 0.002 & 0.494 $\pm$ 0.005 & \textbf{0.503 $\pm$ 0.005} & \textbf{0.495 $\pm$ 0.004} & \textbf{0.642 $\pm$ 0.006} & 0.855 $\pm$ 0.005 \\
M2.2 & {super.} & {SMOTE} & {0.444 $\pm$ 0.004} & {0.364 $\pm$ 0.003} & \textbf{0.521 $\pm$ 0.004} & {0.444 $\pm$ 0.004} & {0.466 $\pm$ 0.004} & 0.596 $\pm$ 0.004 & 0.843 $\pm$ 0.004 \\

\end{tabular}}
    \caption{Results for the decoded-based classification model in combination with different trajectory-level ground truth generation techniques.}
    \label{tab:res_13classifier}
\end{table}

By looking at the variations in weighted accuracy, the models trained on the augmented dataset achieve significantly larger results as compared to the original and imbalanced version of HR-Crime, while the macro accuracy tends to show some decrease. This indicates that the models are focusing on consistently learning an account of all the crime classes when being presented with the same number of trajectory segment samples per class. Comparing M1.2 and M2.2 based on macro accuracy, the model trained on the supervisedly thresholded data (i.e. model M2.2) obtains slightly higher results than the unsupervised based one. In terms of weighted F-score, the results are slightly over $0.4$ score, which indicates that the models are balanced as of precision vs. robustness. 

\begin{figure}[h]
    \centering
     \includegraphics[width=0.67\textwidth]{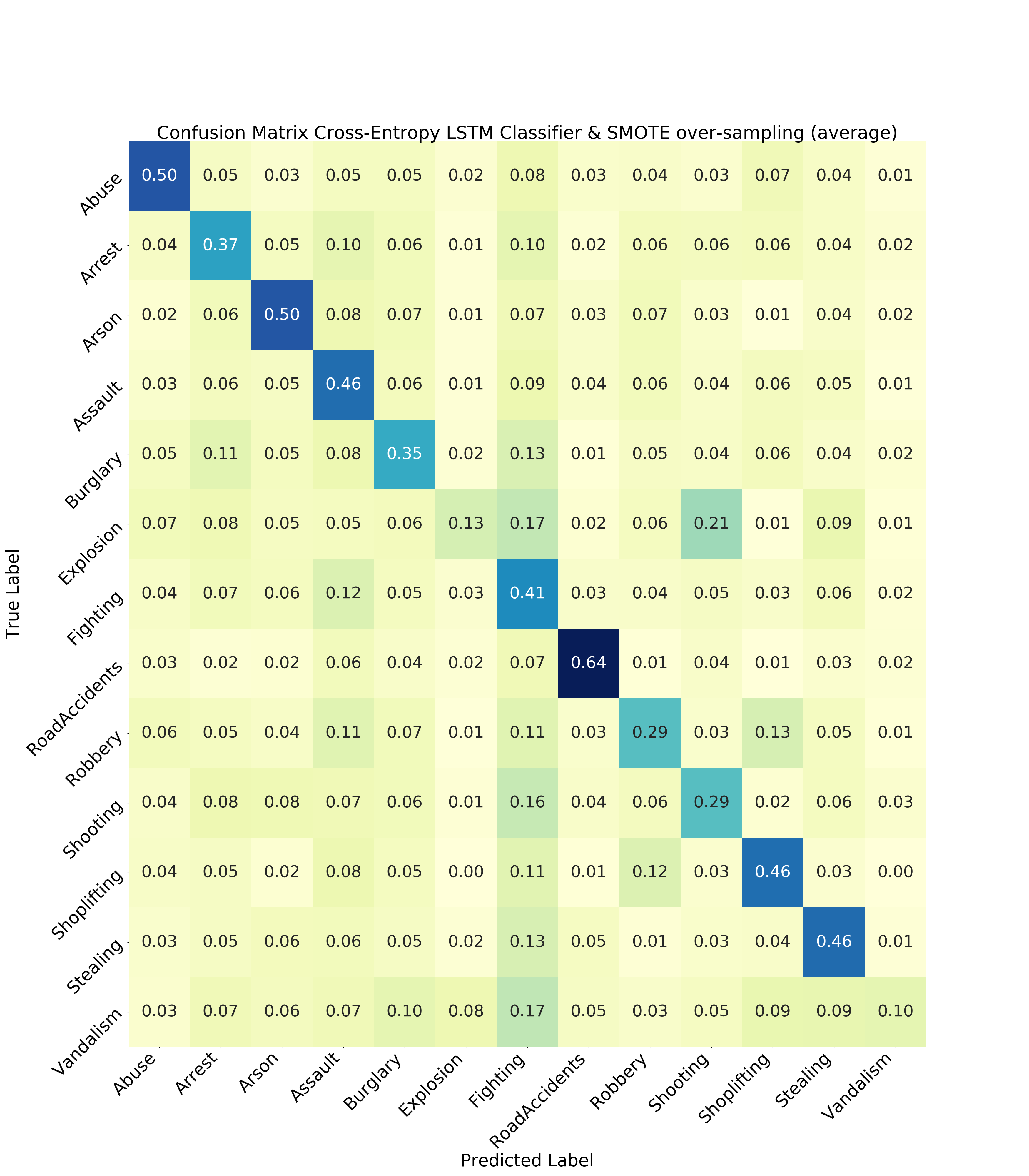}
    \caption{Confusion matrix depicting the average results using M1.2 setting.}
    \label{fig:cm_13class}
\end{figure}

We present the Top-3 and Top-5 macro accuracy for the same experiments presented in Table~\ref{tab:res_13classifier}. The Top-3 accuracy results for all the conducted experiments reach a performance around the $0.6$ score, while for the Top-5 accuracy, the average performance is around the $0.8$ score. The accuracy almost doubles as compared to the Top-1 accuracy scores presented in Table~\ref{tab:res_13classifier}. The model M1.2 achieves $0.577$ in Top-3 and $0.846$ in Top-5 accuracy. This again, indicate low inter-class variability in HR-Crime.

We compare the best performing model M1.2 with the rest of decoded-based classification experiments in Table~\ref{tab:res_13classifier}: M2.2, M1.1, and M2.1. The statistical analysis shows that the obtained results of the models are drawn from a normal distribution, except for the comparison of M1.2 and M2.1. With a confidence threshold $\alpha=0.05$, we apply the null hypothesis stating that the results of the compared models come from the similar distributions. The obtained P-values are 0.056, 0.001, and 0.109, respectively. These values indicate that the null hypothesis can be accepted for 3 of the comparisons. Comparison of our model M1.2 with models M2.2 and M1.1 obtain low P-values which implies that the models do not produce similar results.

For a further quantitative interpretation of the results presented in Table~\ref{tab:res_13classifier}, we analyse the confusion matrix in Figure~\ref{fig:cm_13class} for M1.2 model. The per crime results are consistent with the decrease in overall accuracy noted for MPED-C, as compared to the decoded-based model. Figure~\ref{fig:cm_13class} shows that the most accurate classification is achieved for class \textit{Road Accidents}, with an average accuracy of $0.64$. This is explained by the very distinctive nature of actions captured by the videos in this crime class: the number of skeletons participating in this type of crime is limited since the person driving the vehicle is usually occluded, while the person being affected by the crime (i.e. the accident) is more clearly visible, especially in the first time-steps of the accident. Other classes for which an accuracy performance close to the $0.5$ score is obtained are \textit{Abuse}, \textit{Arson}, \textit{Shoplifting}, and \textit{Stealing}. Surprisingly, there is almost no confusion between classes \textit{Stealing} and \textit{Shoplifting} given the similar nature of the actions involved with these two crimes. This indicates a good demarcation between these two classes. As indicated by the confusion matrix in Figure~\ref{fig:cm_13class}, crimes such as \textit{Explosion}, \textit{Shooting}, and \textit{Vandalism} are often misclassified as class \textit{Fighting}. This situation occurs due to the violent and aggressive nature of actions belonging to this class, which can be quite common to crimes in general. What appears more striking from this confusion matrix is the misinterpretation between class \textit{Explosion} and \textit{Shooting} which occurs 21\% of the time. This might be the case since people have similar fleeing reactions triggered by the loud noise factor that is common for the two events.

The confusion matrix in Figure~\ref{fig:cm_13class} also shows that two crime classes obtain accuracy scores around the $0.1$ score: \textit{Explosion} and \textit{Vandalism}.

We deduce that there exist two influencing factors: low-class representation and feature involvement. The two classes are the least represented in the non-augmented HR-Crime dataset, with 50 trajectories for class \textit{Explosion} and 45 for \textit{Vandalism} (see Table~\ref{tab:dataset}).

Quantitatively comparing the proposed classification model based on decoded trajectory reconstructions and the MPED-C model (see Section \ref{sec:methodology_mped_c}) shows that the decoded-based model (M1.2 model ID) achieves a marginally better performance (MPED-C5.2 model ID). Both models are trained on the proposed unsupervised trajectory-level generated ground truth with an augmented version of 13 crime categories of HR-Crime dataset. The decoded-based model M1.2 is quantitatively superior with an average increase of $0.04$ over all the evaluation metrics. The most significant increase is reported for the weighted accuracy: model M1.2 achieves a $0.38$ score, while the MPED-C5.2 obtains a result of $0.30$. A plausible explanations for the superior performance of the decoded-based model might be regarding the data augmentation techniques applied to each setting. SMOTE, used for model M1.2, is an automatized methodology that infers new data points from neighbouring original samples. The `shifting' technique used for MPED-C5.2 is more laborious and prone to error since the `shifting' factor must be determined empirically. Moreover, it is difficult to generalize the same factor for all samples. Due to some limitations detailed in Section \ref{sec:dataset_augmentation}, the SMOTE technique could not be applied for the case of model MPED-C5.2. Therefore, these incongruencies in data augmentation between the models might explain the performance gap. 

Furthermore, the input data of the decoded-based M1.2 model is the trajectories reconstructions produced by MPED-RNN. Following from the reconstruction process used by MPED-RNN, these representations carry an implicit notion of the anomaly score proposed by MPED-RNN. The MPED-C5.2 model, however, receives the trajectory encodings of MPED-RNN; the encodings are produced by the first component (i.e. encoder) of the MPED-RNN network, at which point the anomaly score is not yet known. This indicates that the MPED-C model is required to infer independently the anomaly characteristics of the trajectories. In this sense, the reconstruction based model M1.2 has an advantage since it implicitly includes that information.

\section{Conclusion and Future work}
\label{sec:conclusion}
 
In this work, we explored the task of human-related crime classification from surveillance footage using the HR-Crime dataset. Given the extracted skeletal trajectories, we proposed a supervised as well as an unsupervised methodology to generate trajectory-level abnormality ground truth and introduced two methods for trajectory-level augmentation of HR-Crime. Later, we present our crime classification framework that is based on human skeleton trajectories and builds on top of the MPED-RNN encoder-decoder architecture for human-related anomaly classification. We present an extensive body of experiments to assess the proposed framework and obtained top-weighted accuracy performances of $0.30$ (encoded-based) and $0.38$ (decoded-based) for the 13 crimes classification. The obtained increase in top-5 accuracy performance reinforced the complexity of the dataset, following from inter-class similarities. For the normal and crime activities classification, the obtained top-weighted accuracy performance of $0.24$  indicated the added challenge posed by the high intra-class variance of the `normal' category. With our experiments we confirmed the negative impact of the dataset imbalance and the suitability of the proposed data augmentation techniques for the task at hand leading to significant boost in classification performance. 

Through our research, we demonstrated that human skeletal trajectory analysis is a feasible approach to crime-related anomaly classification. Possibilities for future work include extending the novel HR-Crime dataset with labels that capture more accurately the nature and boundaries of anomalous events in the surveillance videos. As confirmed by the significantly high top-5 performances of our proposed approach, we expect the multi-label definition of crime categories as well as more accurate definition of non-crime activities to be a sensible direction for future endeavours in the field.

\bibliographystyle{elsarticle-harv}
\bibliography{bibliography.bib}

\begin{thebibliography}{38}
\expandafter\ifx\csname natexlab\endcsname\relax\def\natexlab#1{#1}\fi
\providecommand{\url}[1]{\texttt{#1}}
\providecommand{\href}[2]{#2}
\providecommand{\path}[1]{#1}
\providecommand{\DOIprefix}{doi:}
\providecommand{\ArXivprefix}{arXiv:}
\providecommand{\URLprefix}{URL: }
\providecommand{\Pubmedprefix}{pmid:}
\providecommand{\doi}[1]{\href{http://dx.doi.org/#1}{\path{#1}}}
\providecommand{\Pubmed}[1]{\href{pmid:#1}{\path{#1}}}
\providecommand{\bibinfo}[2]{#2}
\ifx\xfnm\relax \def\xfnm[#1]{\unskip,\space#1}\fi
\bibitem[{Ahmed et~al.(2018)Ahmed, Dogra, Kar and Roy}]{ahmed2018surveillance}
\bibinfo{author}{Ahmed, S.A.}, \bibinfo{author}{Dogra, D.P.},
  \bibinfo{author}{Kar, S.}, \bibinfo{author}{Roy, P.P.}, \bibinfo{year}{2018}.
\newblock \bibinfo{title}{Surveillance scene representation and trajectory
  abnormality detection using aggregation of multiple concepts}.
\newblock \bibinfo{journal}{Expert Systems with Applications}
  \bibinfo{volume}{101}, \bibinfo{pages}{43--55}.
\bibitem[{Akcay et~al.(2018)Akcay, Atapour-Abarghouei and Breckon}]{gans1}
\bibinfo{author}{Akcay, S.}, \bibinfo{author}{Atapour-Abarghouei, A.},
  \bibinfo{author}{Breckon, T.P.}, \bibinfo{year}{2018}.
\newblock \bibinfo{title}{Ganomaly: Semi-supervised anomaly detection via
  adversarial training}.
\newblock \bibinfo{journal}{Asian Conference on Computer Vision} .
\bibitem[{Ak{\c{c}}ay et~al.(2019)Ak{\c{c}}ay, Atapour-Abarghouei and
  Breckon}]{security_screening}
\bibinfo{author}{Ak{\c{c}}ay, S.}, \bibinfo{author}{Atapour-Abarghouei, A.},
  \bibinfo{author}{Breckon, T.P.}, \bibinfo{year}{2019}.
\newblock \bibinfo{title}{Skip-ganomaly: Skip connected and adversarially
  trained encoder-decoder anomaly detection}.
\newblock \bibinfo{journal}{Int. Joint Conf. on Neural Networks} .
\bibitem[{Arroyo et~al.(2015)Arroyo, Yebes, Bergasa, Daza and
  Almaz{\'a}n}]{arroyo2015expert}
\bibinfo{author}{Arroyo, R.}, \bibinfo{author}{Yebes, J.J.},
  \bibinfo{author}{Bergasa, L.M.}, \bibinfo{author}{Daza, I.G.},
  \bibinfo{author}{Almaz{\'a}n, J.}, \bibinfo{year}{2015}.
\newblock \bibinfo{title}{Expert video-surveillance system for real-time
  detection of suspicious behaviors in shopping malls}.
\newblock \bibinfo{journal}{Expert systems with Applications}
  \bibinfo{volume}{42}, \bibinfo{pages}{7991--8005}.
\bibitem[{Boekhoudt et~al.(2021)Boekhoudt, Matei, Aghaei and
  Talavera}]{hr_crime}
\bibinfo{author}{Boekhoudt, K.}, \bibinfo{author}{Matei, A.},
  \bibinfo{author}{Aghaei, M.}, \bibinfo{author}{Talavera, E.},
  \bibinfo{year}{2021}.
\newblock \bibinfo{title}{Hr-crime: Human-related anomaly detection in
  surveillance videos}.
\newblock \bibinfo{journal}{Conf. on Computer Analysis of Images and Patterns}
  .
\bibitem[{Chandola et~al.(2009)Chandola, Banerjee and Kumar}]{anomaly_survey}
\bibinfo{author}{Chandola, V.}, \bibinfo{author}{Banerjee, A.},
  \bibinfo{author}{Kumar, V.}, \bibinfo{year}{2009}.
\newblock \bibinfo{title}{Anomaly detection: A survey}.
\newblock \bibinfo{journal}{ACM Computing Surveys} \bibinfo{volume}{41},
  \bibinfo{pages}{1--58}.
\bibitem[{Chawla et~al.(2002)Chawla, Bowyer, Hall and Kegelmeyer}]{smote}
\bibinfo{author}{Chawla, N.V.}, \bibinfo{author}{Bowyer, K.W.},
  \bibinfo{author}{Hall, L.O.}, \bibinfo{author}{Kegelmeyer, W.P.},
  \bibinfo{year}{2002}.
\newblock \bibinfo{title}{Smote: synthetic minority over-sampling technique}.
\newblock \bibinfo{journal}{Artificial Intelligence Research}
  \bibinfo{volume}{16}, \bibinfo{pages}{321--357}.
\bibitem[{Collins et~al.(2000)Collins, Lipton and
  Kanade}]{intro_automatic_surveillance}
\bibinfo{author}{Collins, R.T.}, \bibinfo{author}{Lipton, A.J.},
  \bibinfo{author}{Kanade, T.}, \bibinfo{year}{2000}.
\newblock \bibinfo{title}{Introduction to the special section on video
  surveillance}.
\newblock \bibinfo{journal}{Transactions on Pattern Analysis and Machine
  Intelligence} \bibinfo{volume}{22}, \bibinfo{pages}{745--746}.
\bibitem[{Dalal and Triggs(2005)}]{oriented_gradients}
\bibinfo{author}{Dalal, N.}, \bibinfo{author}{Triggs, B.},
  \bibinfo{year}{2005}.
\newblock \bibinfo{title}{Histograms of oriented gradients for human
  detection}.
\newblock \bibinfo{journal}{Conf. on Computer Vision and Pattern Recognition} .
\bibitem[{Doshi and Yilmaz(2020)}]{any_shot}
\bibinfo{author}{Doshi, K.}, \bibinfo{author}{Yilmaz, Y.},
  \bibinfo{year}{2020}.
\newblock \bibinfo{title}{Any-shot sequential anomaly detection in surveillance
  videos}.
\newblock \bibinfo{journal}{Conf. on Computer Vision and Pattern Recognition} .
\bibitem[{Doshi and Yilmaz(2021)}]{doshi2021online}
\bibinfo{author}{Doshi, K.}, \bibinfo{author}{Yilmaz, Y.},
  \bibinfo{year}{2021}.
\newblock \bibinfo{title}{Online anomaly detection in surveillance videos with
  asymptotic bound on false alarm rate}.
\newblock \bibinfo{journal}{Pattern Recognition} , \bibinfo{pages}{107865}.
\bibitem[{Emonet et~al.(2011)Emonet, Varadarajan and Odobez}]{human_anomaly}
\bibinfo{author}{Emonet, R.}, \bibinfo{author}{Varadarajan, J.},
  \bibinfo{author}{Odobez, J.M.}, \bibinfo{year}{2011}.
\newblock \bibinfo{title}{Multi-camera open space human activity discovery for
  anomaly detection}.
\newblock \bibinfo{journal}{Conf. on Advanced Video and Signal Based
  Surveillance} .
\bibitem[{Febin et~al.(2020)Febin, Jayasree and Joy}]{violence1}
\bibinfo{author}{Febin, I.}, \bibinfo{author}{Jayasree, K.},
  \bibinfo{author}{Joy, P.T.}, \bibinfo{year}{2020}.
\newblock \bibinfo{title}{Violence detection in videos for an intelligent
  surveillance system using mobsift and movement filtering algorithm}.
\newblock \bibinfo{journal}{Pattern Analysis and Applications}
  \bibinfo{volume}{23}, \bibinfo{pages}{611--623}.
\bibitem[{Gehan(1965)}]{gehan1965generalized}
\bibinfo{author}{Gehan, E.A.}, \bibinfo{year}{1965}.
\newblock \bibinfo{title}{A generalized wilcoxon test for comparing arbitrarily
  singly-censored samples}.
\newblock \bibinfo{journal}{Biometrika} \bibinfo{volume}{52},
  \bibinfo{pages}{203--223}.
\bibitem[{Greff et~al.(2016)Greff, Srivastava, Koutn{\'\i}k, Steunebrink and
  Schmidhuber}]{lstm}
\bibinfo{author}{Greff, K.}, \bibinfo{author}{Srivastava, R.K.},
  \bibinfo{author}{Koutn{\'\i}k, J.}, \bibinfo{author}{Steunebrink, B.R.},
  \bibinfo{author}{Schmidhuber, J.}, \bibinfo{year}{2016}.
\newblock \bibinfo{title}{Lstm: A search space odyssey}.
\newblock \bibinfo{journal}{Transactions on Neural Networks and Learning
  Systems} \bibinfo{volume}{28}, \bibinfo{pages}{2222--2232}.
\bibitem[{Hasan et~al.(2016)Hasan, Choi, Neumann, Roy-Chowdhury and
  Davis}]{deeplearning1}
\bibinfo{author}{Hasan, M.}, \bibinfo{author}{Choi, J.},
  \bibinfo{author}{Neumann, J.}, \bibinfo{author}{Roy-Chowdhury, A.K.},
  \bibinfo{author}{Davis, L.S.}, \bibinfo{year}{2016}.
\newblock \bibinfo{title}{Learning temporal regularity in video sequences}.
\newblock \bibinfo{journal}{Conf. on Computer Vision and Pattern Recognition} .
\bibitem[{Hsu and Lachenbruch(2005)}]{hsu2005paired}
\bibinfo{author}{Hsu, H.}, \bibinfo{author}{Lachenbruch, P.A.},
  \bibinfo{year}{2005}.
\newblock \bibinfo{title}{Paired t test}.
\newblock \bibinfo{journal}{Encyclopedia of Biostatistics} \bibinfo{volume}{6}.
\bibitem[{Kingma and Ba(2014)}]{adam}
\bibinfo{author}{Kingma, D.P.}, \bibinfo{author}{Ba, J.}, \bibinfo{year}{2014}.
\newblock \bibinfo{title}{Adam: A method for stochastic optimization}.
\newblock \bibinfo{journal}{Conf. on Learning Representations} .
\bibitem[{Lee et~al.(2018)Lee, Leong, Lai, Leow and Yap}]{lee2018archcam}
\bibinfo{author}{Lee, W.K.}, \bibinfo{author}{Leong, C.F.},
  \bibinfo{author}{Lai, W.K.}, \bibinfo{author}{Leow, L.K.},
  \bibinfo{author}{Yap, T.H.}, \bibinfo{year}{2018}.
\newblock \bibinfo{title}{Archcam: Real time expert system for suspicious
  behaviour detection in atm site}.
\newblock \bibinfo{journal}{Expert Systems with Applications}
  \bibinfo{volume}{109}, \bibinfo{pages}{12--24}.
\bibitem[{Lin and Purnell(2019)}]{billion_cameras}
\bibinfo{author}{Lin, L.}, \bibinfo{author}{Purnell, N.}, \bibinfo{year}{2019}.
\newblock \bibinfo{title}{A world with a billion cameras watching you is just
  around the corner}.
\newblock \bibinfo{journal}{The Wall Street Journal} .
\bibitem[{Loganathan et~al.(2019)Loganathan, Kariyawasam and
  Sumathipala}]{crime}
\bibinfo{author}{Loganathan, S.}, \bibinfo{author}{Kariyawasam, G.},
  \bibinfo{author}{Sumathipala, P.}, \bibinfo{year}{2019}.
\newblock \bibinfo{title}{Suspicious activity detection in surveillance
  footage}.
\newblock \bibinfo{journal}{Conf. on Electrical and Computing Technologies and
  Applications} .
\bibitem[{Luo et~al.(2017)Luo, Liu and Gao}]{sparsecodingrnn}
\bibinfo{author}{Luo, W.}, \bibinfo{author}{Liu, W.}, \bibinfo{author}{Gao,
  S.}, \bibinfo{year}{2017}.
\newblock \bibinfo{title}{A revisit of sparse coding based anomaly detection in
  stacked rnn framework}.
\newblock \bibinfo{journal}{Conf. on Computer Vision} .
\bibitem[{Malhotra et~al.(2016)Malhotra, Ramakrishnan, Anand, Vig, Agarwal and
  Shroff}]{enc_dec_multi_sensor}
\bibinfo{author}{Malhotra, P.}, \bibinfo{author}{Ramakrishnan, A.},
  \bibinfo{author}{Anand, G.}, \bibinfo{author}{Vig, L.},
  \bibinfo{author}{Agarwal, P.}, \bibinfo{author}{Shroff, G.},
  \bibinfo{year}{2016}.
\newblock \bibinfo{title}{Lstm-based encoder-decoder for multi-sensor anomaly
  detection}.
\newblock \bibinfo{journal}{Conf. on Machine Learning} .
\bibitem[{Markou and Singh(2003)}]{novelty_1}
\bibinfo{author}{Markou, M.}, \bibinfo{author}{Singh, S.},
  \bibinfo{year}{2003}.
\newblock \bibinfo{title}{Novelty detection: A review—part 1: Statistical
  approaches}.
\newblock \bibinfo{journal}{Signal Processing} \bibinfo{volume}{83},
  \bibinfo{pages}{2481--2497}.
\bibitem[{Maron and Lozano-P{\'e}rez(1998)}]{mil}
\bibinfo{author}{Maron, O.}, \bibinfo{author}{Lozano-P{\'e}rez, T.},
  \bibinfo{year}{1998}.
\newblock \bibinfo{title}{A framework for multiple-instance learning}.
\newblock \bibinfo{journal}{Advances in Neural Information Processing Systems}
  .
\bibitem[{Morais et~al.(2019)Morais, Le, Tran, Saha, Mansour and
  Venkatesh}]{mpedrnn}
\bibinfo{author}{Morais, R.}, \bibinfo{author}{Le, V.}, \bibinfo{author}{Tran,
  T.}, \bibinfo{author}{Saha, B.}, \bibinfo{author}{Mansour, M.},
  \bibinfo{author}{Venkatesh, S.}, \bibinfo{year}{2019}.
\newblock \bibinfo{title}{Learning regularity in skeleton trajectories for
  anomaly detection in videos}.
\newblock \bibinfo{journal}{Conf. on Computer Vision and Pattern Recognition} .
\bibitem[{Nasaruddin et~al.(2020)Nasaruddin, Muchtar, Afdhal and
  Dwiyantoro}]{visual_attention}
\bibinfo{author}{Nasaruddin, N.}, \bibinfo{author}{Muchtar, K.},
  \bibinfo{author}{Afdhal, A.}, \bibinfo{author}{Dwiyantoro, A.P.J.},
  \bibinfo{year}{2020}.
\newblock \bibinfo{title}{Deep anomaly detection through visual attention in
  surveillance videos}.
\newblock \bibinfo{journal}{Journal of Big Data} \bibinfo{volume}{7},
  \bibinfo{pages}{1--17}.
\bibitem[{Rabie(1990)}]{rabie1990applications}
\bibinfo{author}{Rabie, S.}, \bibinfo{year}{1990}.
\newblock \bibinfo{title}{Applications of expert systems to network
  surveillance}, in: \bibinfo{booktitle}{Network Management and Control}.
  \bibinfo{publisher}{Springer}, pp. \bibinfo{pages}{249--262}.
\bibitem[{Ramachandra et~al.(2020)Ramachandra, Jones and
  Vatsavai}]{video_anomaly_detection_survey}
\bibinfo{author}{Ramachandra, B.}, \bibinfo{author}{Jones, M.},
  \bibinfo{author}{Vatsavai, R.R.}, \bibinfo{year}{2020}.
\newblock \bibinfo{title}{A survey of single-scene video anomaly detection}.
\newblock \bibinfo{journal}{Transactions on Pattern Analysis and Machine
  Intelligence} .
\bibitem[{Reynolds(2009)}]{gmm}
\bibinfo{author}{Reynolds, D.A.}, \bibinfo{year}{2009}.
\newblock \bibinfo{title}{Gaussian mixture models}.
\newblock \bibinfo{journal}{Encyclopedia of Biometrics} \bibinfo{volume}{741},
  \bibinfo{pages}{659--663}.
\bibitem[{Shapiro and Wilk(1965)}]{SHAPIRO1965}
\bibinfo{author}{Shapiro, S.S.}, \bibinfo{author}{Wilk, M.B.},
  \bibinfo{year}{1965}.
\newblock \bibinfo{title}{An analysis of variance test for normality (complete
  samples)}.
\newblock \bibinfo{journal}{Biometrika} \bibinfo{volume}{52},
  \bibinfo{pages}{591--611}.
\bibitem[{Sultani et~al.(2018a)Sultani, Chen and Shah}]{ucf_crime}
\bibinfo{author}{Sultani, W.}, \bibinfo{author}{Chen, C.},
  \bibinfo{author}{Shah, M.}, \bibinfo{year}{2018}a.
\newblock \bibinfo{title}{Real-world anomaly detection in surveillance videos}.
\newblock \bibinfo{journal}{Conf. on Computer Vision and Pattern Recognition} .
\bibitem[{Sultani et~al.(2018b)Sultani, Chen and Shah}]{real_world}
\bibinfo{author}{Sultani, W.}, \bibinfo{author}{Chen, C.},
  \bibinfo{author}{Shah, M.}, \bibinfo{year}{2018}b.
\newblock \bibinfo{title}{Real-world anomaly detection in surveillance videos}.
\newblock \bibinfo{journal}{Conf. on Computer Vision and Pattern Recognition} .
\bibitem[{Tran et~al.(2015)Tran, Bourdev, Fergus, Torresani and
  Paluri}]{c3d_feat}
\bibinfo{author}{Tran, D.}, \bibinfo{author}{Bourdev, L.},
  \bibinfo{author}{Fergus, R.}, \bibinfo{author}{Torresani, L.},
  \bibinfo{author}{Paluri, M.}, \bibinfo{year}{2015}.
\newblock \bibinfo{title}{Learning spatiotemporal features with 3d
  convolutional networks}.
\newblock \bibinfo{journal}{Conf. on Computer Vision} .
\bibitem[{Ullah et~al.(2021)Ullah, Ullah, Haq, Muhammad, Sajjad and
  Baik}]{lstmcnn}
\bibinfo{author}{Ullah, W.}, \bibinfo{author}{Ullah, A.}, \bibinfo{author}{Haq,
  I.}, \bibinfo{author}{Muhammad, K.}, \bibinfo{author}{Sajjad, M.},
  \bibinfo{author}{Baik, S.W.}, \bibinfo{year}{2021}.
\newblock \bibinfo{title}{Cnn features with bi-directional lstm for real-time
  anomaly detection in surveillance networks}.
\newblock \bibinfo{journal}{Multimedia Tools and Applications}
  \bibinfo{volume}{80}, \bibinfo{pages}{16979--16995}.
\bibitem[{Yun et~al.(2014)Yun, Jeong, Yi, Kim and Choi}]{car1}
\bibinfo{author}{Yun, K.}, \bibinfo{author}{Jeong, H.}, \bibinfo{author}{Yi,
  K.M.}, \bibinfo{author}{Kim, S.W.}, \bibinfo{author}{Choi, J.Y.},
  \bibinfo{year}{2014}.
\newblock \bibinfo{title}{Motion interaction field for accident detection in
  traffic surveillance video}.
\newblock \bibinfo{journal}{Conf. on Pattern Recognition} .
\bibitem[{Zhang and Sabuncu(2018)}]{cross_entropy}
\bibinfo{author}{Zhang, Z.}, \bibinfo{author}{Sabuncu, M.R.},
  \bibinfo{year}{2018}.
\newblock \bibinfo{title}{Generalized cross entropy loss for training deep
  neural networks with noisy labels}.
\newblock \bibinfo{journal}{Conf. on Neural Information Processing Systems} .
\bibitem[{Zhou and Kwan(2018)}]{opticalflow1}
\bibinfo{author}{Zhou, J.}, \bibinfo{author}{Kwan, C.}, \bibinfo{year}{2018}.
\newblock \bibinfo{title}{Anomaly detection in low quality traffic monitoring
  videos using optical flow}.
\newblock \bibinfo{journal}{Pattern Recognition and Tracking XXIX}
  \bibinfo{volume}{10649}, \bibinfo{pages}{122--132}.

\end{thebibliography}

\end{document}